\documentclass[preprint,12pt]{elsarticle}
\usepackage{amsfonts}
\usepackage{amsmath}
\usepackage{xcolor}
\usepackage{amsthm}
\usepackage{multirow}
\usepackage{booktabs}
\usepackage{graphicx}
\usepackage{subcaption}
\usepackage{enumitem}
\usepackage{float}
\usepackage{url}
\usepackage{longtable}
\usepackage{caption}
\usepackage{hyperref}
\usepackage{bm}
\usepackage{placeins} 
\begin{document}

\begin{frontmatter}

\title{Relative Discharge Stage Classification from Battery Sensor Measurements in Laboratory Discharge Tests}

\author[1]{Khoa~Tran}
\ead{trandinhkhoa@tdtu.edu.vn}

\author[2]{Tri~Le}
\ead{tri.le@aiware.website}

\author[3]{Hung-Cuong Trinh}
\ead{trinhhungcuong@tdtu.edu.vn}

\author[4,5]{Hung Tran-Nam\corref{cor1}}
\ead{hung.trannam@vlu.edu.vn}

\cortext[cor1]{Corresponding author: \href{mailto:hung.trannam@vlu.edu.vn}{hung.trannam@vlu.edu.vn}}

\affiliation[1]{organization={Data Science Laboratory, Faculty of Information Technology, Ton Duc Thang University},  
addressline={Ho Chi Minh City, Vietnam},  
city={Ho Chi Minh City},  
postcode={70000},  
country={Vietnam}}

\affiliation[2]{organization={AIWARE Limited Company},  
addressline={17 Huynh Man Dat Street, Hoa Cuong Ward, Hai Chau District},  
city={Da Nang},  
postcode={550000},  
country={Vietnam}}  
 
\affiliation[3]{organization={Natural Language Processing and Knowledge Discovery Research Group, Faculty of Information Technology, Ton Duc Thang University},  
city={Ho Chi Minh City},  
postcode={70000},  
country={Vietnam}}  

\affiliation[4]{organization={Laboratory for Applied and Industrial Mathematics, Institute for Computational Science and Artificial Intelligence, Van Lang University},  
city={Ho Chi Minh City},  
postcode={70000},  
country={Vietnam}}  

\affiliation[5]{organization={Faculty of Basic Sciences, Van Lang University},  
city={Ho Chi Minh City},  
postcode={70000},  
country={Vietnam}}  


\begin{abstract}
Relative discharge stage (RDS) is an ordinal target obtained by dividing the realized remaining discharge time by the total duration of the same discharge episode and assigning the ratio to a bin. We investigate whether these bins can be predicted from measurements available up to the present sample. The framework combines an electrothermal equivalent-circuit model and adaptive extended Kalman filter to estimate state of charge (SOC), then classifies discharge stage using the estimated SOC, current, terminal voltage, and surface temperature. With the evaluated two-layer convolutional network, the final-time prediction directly uses the three most recent observations. Earlier measurements can influence the output through the recursive SOC estimate. On complete discharge tests, the five-stage classifier achieved accuracies of \(95.38\%\) and \(81.23\%\) for FTP-75 and PDMHC and \(76.41\%\) and \(73.87\%\) for LCO\_5 and LCO\_6, respectively. These results characterize classification within the laboratory protocols. Incremental value over SOC alone, reference-based SOC accuracy, and utility in vehicle operation remain to be established.
\end{abstract}


\end{frontmatter}

\section{Introduction}

Lithium-ion batteries are widely used in electric vehicles (EVs) and other 
electrified applications owing to their high energy density, long cycle 
life, low self-discharge rate, and reliability~\cite{kim2019lithium}. 
The increasing adoption of electric transportation has further emphasized 
the importance of reliable battery monitoring and management
~\cite{timilsina2025global}. To support safe and efficient battery operation, 
battery management systems (BMSs) utilize measured signals, such as current, 
terminal voltage, and temperature, to estimate battery conditions 
~\cite{gabbar2021review}. These estimation tasks include life-cycle 
prognostics, such as state of health (SOH)
~\cite{ren2026physics,kham2026bayesian} and remaining useful life (RUL)
~\cite{li2026joint,peng2026novel}, as well as in-cycle estimation, such as 
state of charge (SOC)
~\cite{wu2026equivalent,zheng2026lithium,jiao2026methodology,shah2026review} 
and remaining discharge time (RDT)~\cite{wang2020framework}. While 
life-cycle prognostics characterize long-term battery degradation, 
in-cycle estimation focuses on the battery's current condition and 
remaining discharge capability.

SOC is one of the most widely studied battery states because it 
characterizes the amount of charge remaining relative to a reference 
capacity. To support deployment on embedded devices, 
\cite{li2025lightweight} proposed a lightweight SOC estimation method 
that combines Savitzky--Golay filtering, BOHB-based hyperparameter 
optimization, and neural-network pruning. To address variations in 
battery chemistry, ambient temperature, and the scarcity of labeled 
training data, \cite{badfar2025state} proposed a regression-based 
unsupervised domain adaptation framework that transfers knowledge from 
a labeled source battery to an unlabeled target battery. Although SOC 
provides essential information about the current charge condition, 
it does not directly specify how long a battery can continue operating 
before reaching its discharge cutoff. The remaining operating duration 
also depends on future load demand, temperature, battery characteristics, 
and the prescribed cutoff condition. This has motivated research on 
RDT and related quantities, such as remaining discharge energy (RDE).

Existing RDT estimation methods can generally be classified into two 
categories: (1) methods that assume future current profiles are known 
and (2) methods that operate without direct knowledge of future current 
profiles. In the first category, \cite{daigle2016end} employed an 
electrochemistry-based battery model combined with an unscented Kalman 
filter (UKF)~\cite{wan2000unscented} to estimate internal battery states 
from measured current and voltage. Given the estimated states and a 
specified future current profile, the model predicts future terminal 
voltage until the discharge cutoff is reached. The remaining discharge 
time is then calculated as
\begin{equation}
    \mathrm{RDT}(t)=t_{\mathrm{EOD}}-t,
    \label{eq:intro_rdt}
\end{equation}
where \(t_{\mathrm{EOD}}\) denotes the end-of-discharge (EOD) time.

Similarly, \cite{chen2019particle} employed a particle filter to 
estimate open-circuit voltage (OCV) and determine SOC through the 
OCV--SOC relationship. Given a future current profile, the SOC 
trajectory is propagated using Coulomb counting, while an equivalent 
circuit model (ECM) predicts the terminal voltage until the cutoff 
condition is reached. Such model-based approaches may incur substantial 
computational costs because battery states must be propagated over the 
remaining discharge horizon. In related work, \cite{tu2024remaining} 
investigated feedforward neural networks for direct remaining-discharge 
prediction using internal battery states and specified future operating 
conditions, thereby reducing the need for repeated forward simulation. 
Nevertheless, methods that rely on known future operating conditions 
remain limited when future load demand is unavailable or uncertain.

To address the absence of known future current profiles, several 
studies have incorporated load-prediction mechanisms into 
remaining-discharge estimation frameworks. For example, 
\cite{wang2020framework} employed a discrete wavelet transform 
(DWT)-based current-prediction method combined with an RC circuit model 
and an unscented particle filter (UPF) to estimate future terminal 
voltage and determine the EOD time. In the context of RDE estimation, 
\cite{hatherall2023remaining} combined driving-pattern recognition 
with Markov-based future load prediction. Driving patterns were 
identified using principal component analysis and K-means clustering, 
while the predicted load sequence was subsequently used in a 
first-order ECM. Similarly, \cite{lai2022remaining} investigated 
hidden Markov model (HMM)-based future current prediction to estimate 
remaining discharge capability while accounting for operating 
conditions. These approaches explicitly predict future load profiles 
before estimating the remaining discharge characteristics. Consequently, 
their performance depends on both the accuracy of the battery model 
and the accuracy of the predicted future loads.

Rather than explicitly forecasting future current or regressing an 
absolute remaining discharge time, this study investigates an ordinal 
formulation referred to as Relative Discharge Stage (RDS). RDS is 
defined by normalizing the realized remaining discharge time by the 
total duration of the corresponding discharge episode and subsequently 
discretizing the resulting ratio into ordered intervals. Specifically, 
the normalized remaining discharge ratio is expressed as
\begin{equation}
    q_t =
    \frac{t_{\mathrm{EOD}}-t}
         {t_{\mathrm{EOD}}-t_{\mathrm{start}}},
    \qquad q_t\in[0,1],
    \label{eq:intro_normalized_rdt}
\end{equation}
where \(t_{\mathrm{start}}\) denotes the beginning of the discharge 
episode. For a classification problem with \(K\) equally spaced 
intervals, the corresponding RDS label is defined as
\begin{equation}
    y_t =
    \min\left(K-1,\left\lfloor Kq_t\right\rfloor\right).
    \label{eq:intro_rds}
\end{equation}

Unlike SOC, which characterizes the remaining charge, normalized RDT 
characterizes the remaining fraction of the realized discharge 
duration. However, these two quantities are closely related under 
certain operating conditions. For an ideal constant-current discharge 
that begins at full SOC and ends at zero SOC, normalized RDT is 
mathematically equivalent to SOC. More generally, when SOC decreases 
linearly throughout a discharge episode, their relationship can be 
expressed as
\begin{equation}
    q_t =
    \frac{S_t-S_{\mathrm{EOD}}}
         {S_{\mathrm{start}}-S_{\mathrm{EOD}}},
    \label{eq:intro_soc_rds_relationship}
\end{equation}
where \(S_t\), \(S_{\mathrm{start}}\), and \(S_{\mathrm{EOD}}\) denote 
the SOC at the current sampling instant, the beginning of discharge, 
and the EOD condition, respectively. Under variable-current operation 
or voltage-dependent cutoff conditions, this relationship does not 
necessarily hold because charge depletion and elapsed discharge time 
may evolve differently. This distinction motivates investigating whether 
measured operating signals can provide complementary information to 
SOC when estimating relative discharge progress.

\begin{figure}[H]
    \centering
    \includegraphics[width=1\textwidth]
    {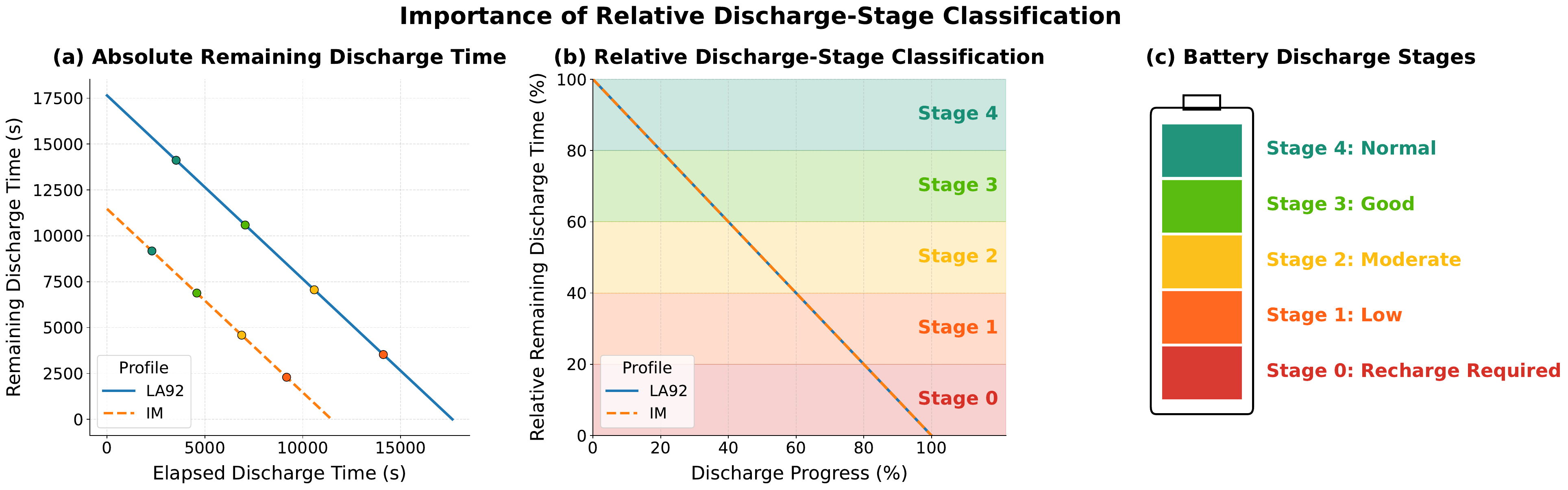}
    \caption{Illustration of relative discharge-stage classification 
    using two discharge profiles with different absolute durations. 
    Normalization maps the realized remaining discharge times onto 
    a common relative scale.}
    \label{fig:rds_concept}
\end{figure}

Figure~\ref{fig:rds_concept} illustrates the motivation for the RDS 
formulation. Discharge episodes with different absolute durations are 
represented on a common relative scale and divided into five ordered 
stages. This representation replaces precise time prediction with a 
coarser classification target. Nevertheless, RDS is a discretized 
representation of normalized RDT rather than an independent physical 
battery state. Discretization changes the granularity of the prediction 
task but does not eliminate the uncertainty associated with unknown 
future loads. Moreover, identical RDS classes do not necessarily 
correspond to identical remaining operating times or operational risks.

The complete discharge duration and EOD time are required only during 
offline label construction. During online inference, the classification 
model receives measurements available up to the current sampling instant 
without access to the realized future current profile or the final 
discharge duration. Consequently, the objective is to investigate the 
extent to which relative discharge stages can be inferred from 
observable battery conditions under the evaluated discharge protocols.

From an engineering perspective, relative discharge-stage classification 
may provide a complementary representation of discharge progress in 
laboratory battery characterization or equipment operating under 
specified duty regimes. For example, a stage-based representation could 
potentially support progress monitoring during discharge tests with 
different absolute durations. However, its practical benefit depends 
on whether it provides useful information beyond conventional SOC 
estimation. In particular, applications involving recharge decisions, 
remaining driving range, or early warnings of battery depletion require 
additional validation under representative operating conditions. 
Real-world EV operation involves partial discharge, regenerative 
braking, intermittent charging, and unpredictable driving demands, 
which complicate the definition and prediction of the discharge 
endpoint. Therefore, this study focuses on complete laboratory 
discharge episodes rather than claiming direct applicability to 
real-vehicle operation.

To investigate this problem, we develop a hybrid framework that 
combines a recursive electrothermal SOC estimator with a lightweight 
temporal classifier. The estimator integrates physical battery dynamics 
and neural residual-voltage correction to construct an estimated SOC 
feature from measured current, terminal voltage, and surface 
temperature. The estimated SOC is subsequently combined with these 
measured signals to classify the relative discharge stage. The main 
methodological contributions of this work are summarized as follows:

\begin{itemize}

    \item \textbf{An ordinal formulation of remaining discharge time:}
    We formulate relative discharge-stage classification as the 
    prediction of discretized normalized RDT from measurements 
    available up to the current sampling instant. This formulation 
    provides a common relative scale for complete discharge episodes 
    with different absolute durations. The relationship between RDS 
    and SOC is explicitly discussed, together with the limitations 
    arising from unknown future operating conditions.

    \item \textbf{A hybrid electrothermal SOC estimation framework:}
    We develop a recursive SOC estimation framework that integrates 
    a second-order ECM, voltage hysteresis, temperature-dependent OCV 
    correction, core-temperature estimation, and an Adaptive Extended 
    Kalman Filter (AEKF)~\cite{xiong2012evaluation}. The framework 
    additionally incorporates an SOC--Temperature-Coupled Equivalent 
    Circuit Model (STC-ECM) parameter adaptation mechanism and a 
    pretrained lightweight neural network for residual-voltage 
    correction. These components are designed to improve 
    terminal-voltage modeling under varying load and thermal 
    conditions while providing an estimated SOC feature for 
    subsequent RDS classification.

    \item \textbf{A physics-informed RDS classification framework:}
    The measured current, terminal voltage, and surface temperature 
    are combined with the recursively estimated SOC and processed 
    using a lightweight Temporal Convolutional Network 
    (TCN)~\cite{lea2017temporal}. The evaluated two-layer TCN 
    directly extracts local temporal information from the three 
    most recent observations, while earlier measurements can 
    influence the classification through the recursive SOC estimate. 
    The classifier predicts one of five ordered discharge stages 
    without requiring explicit future-current prediction during 
    inference.

    \item \textbf{Experimental evaluation across laboratory discharge 
    conditions:}
    The proposed framework is evaluated using complete discharge 
    episodes from two public lithium-ion battery datasets covering 
    NMC and LCO chemistries, different load conditions, and multiple 
    ambient temperatures. The experiments examine terminal-voltage 
    modeling performance, input-feature selection, classifier 
    architecture, and relative discharge-stage classification 
    performance. The results characterize the behavior of the 
    framework under the evaluated laboratory protocols, while 
    identifying the need for further validation of its incremental 
    value beyond SOC, SOC estimation accuracy, and practical 
    engineering utility.

\end{itemize}

The remainder of this paper is organized as follows. 
Section~\ref{Proposal} presents the proposed methodology and 
formulates the RDS classification problem. 
Section~\ref{Experimental_Setup} describes the datasets, experimental 
configurations, baseline methods, and evaluation metrics. 
Section~\ref{Experiments_Results} presents the experimental results. 
Section~\ref{Discussion} discusses the findings, limitations, and 
potential engineering applications of the proposed framework. 
Finally, Section~\ref{Conclusion} concludes the paper and outlines 
directions for future research.

\section{Proposed Method}
\label{Proposal}

\subsection{Overall Framework and RDS Formulation}
\label{subsec:overall_framework}

\begin{figure}[H]
\centering
\includegraphics[width=\textwidth]
{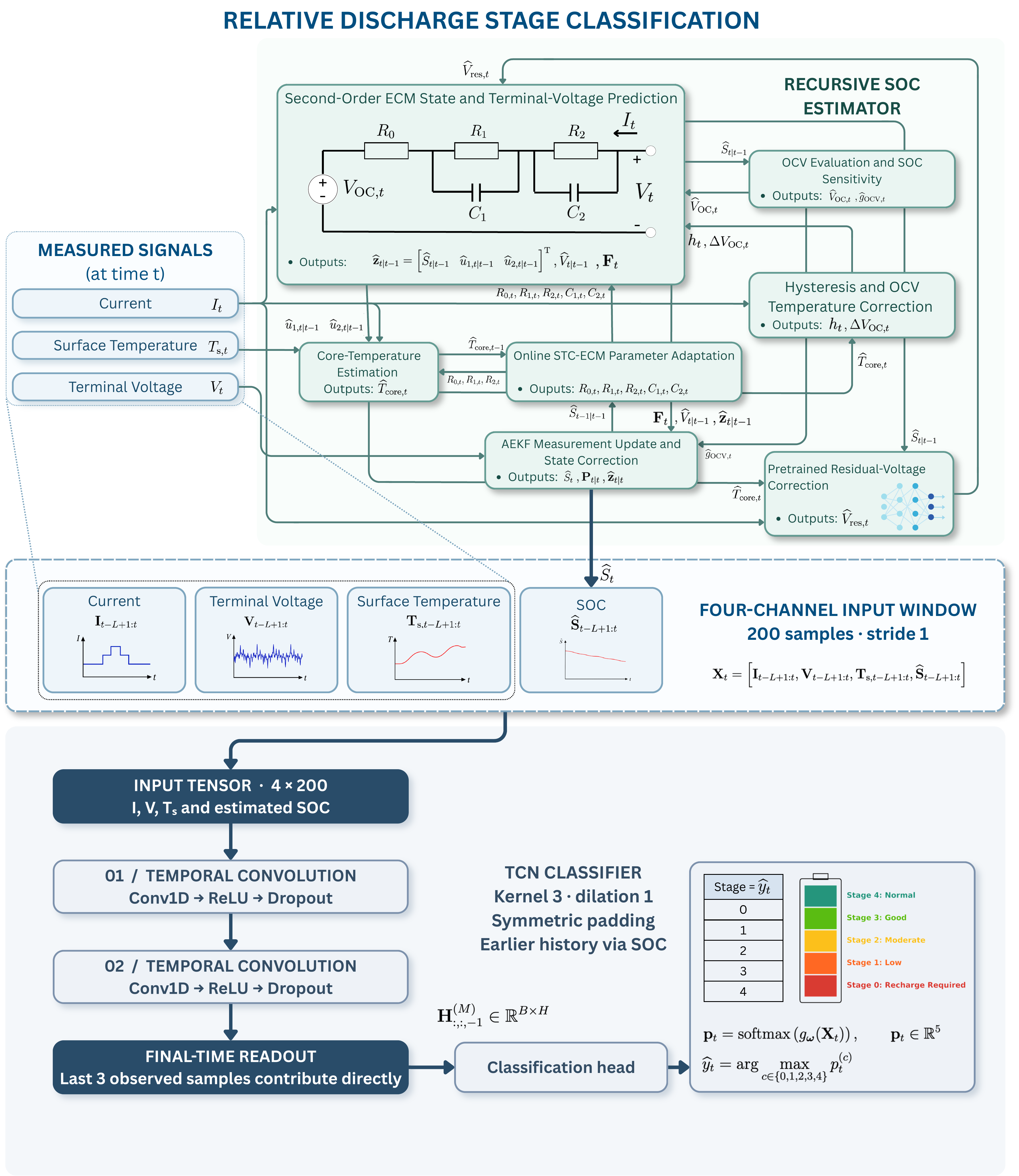}
\caption{Overall framework for relative discharge stage (RDS) classification. Complete trajectories and voltage cutoff times define the labels offline. During online inference, measurements enter both the recursive SOC estimator and the classifier input. The final-time TCN readout directly depends on three observed samples, with earlier information carried through the estimator state.}
\label{fig:RDS_overall_architecture}
\end{figure}

\paragraph{Online Inference Framework}
Figure~\ref{fig:RDS_overall_architecture} illustrates the online inference framework of the proposed RDS classification model. At each sampling instant \(t\), the measured current \(I_t\), terminal
voltage \(V_t\), and surface temperature \(T_{\mathrm{s},t}\) are
processed by the proposed online SOC estimation to estimate the battery SOC. The SOC estimation method is represented as
\begin{equation}
\widehat{S}_t
=
\mathcal{F}_{\boldsymbol{\psi}}
\left(
I_t,
V_t,
T_{\mathrm{s},t},
\widehat{\mathbf{z}}_{t-1|t-1}
\right),
\label{eq:overall_soc_estimator}
\end{equation}
where \(\mathcal{F}_{\boldsymbol{\psi}}(\cdot)\) denotes the proposed
SOC estimation method and \(\widehat{\mathbf{z}}_{t-1|t-1}\) is the posterior
battery-state estimate from the preceding sampling instant.

A physics-informed observation window of length \(L\) is constructed from the measured current, terminal-voltage, and surface-temperature sequences together with the estimated SOC sequence:
\begin{equation}
\mathbf{X}_t
=
\left[
\mathbf{I}_{t-L+1:t},
\mathbf{V}_{t-L+1:t},
\mathbf{T}_{\mathrm{s},t-L+1:t},
\widehat{\mathbf{S}}_{t-L+1:t}
\right].
\label{eq:overall_observation_window}
\end{equation}
Therefore, consecutive windows overlap by \(L-1\) sampling instants, corresponding to a sliding-window stride of one.

The online RDS classifier maps the formatted window to softmax scores over \(K=5\) RDS classes:
\begin{equation}
\mathbf{p}_t
=
\operatorname{softmax}
\left(
g_{\boldsymbol{\omega}}
\left(
\mathbf{X}_t
\right)
\right),
\qquad
\mathbf{p}_t\in\mathbb{R}^{5},
\label{eq:overall_rds_probability}
\end{equation}
where \(p_t^{(c)}\) is the predicted probability of class \(c\). The
predicted RDS class is
\begin{equation}
\widehat{y}_t
=
\underset{c\in\{0,1,2,3,4\}}{\arg\max}
\;
p_t^{(c)}.
\label{eq:overall_rds_prediction}
\end{equation}

\paragraph{Offline RDS Label Construction}
For a uniformly sampled complete discharge episode beginning at index \(N_{\mathrm{start}}\) and ending at the protocol-defined terminal-voltage cutoff at \(N_{\mathrm{EOD}}>N_{\mathrm{start}}\), the relative remaining discharge ratio at sample \(t\) is defined as
\begin{equation}
q_t
=
\frac{
N_{\mathrm{EOD}}-t
}{
N_{\mathrm{EOD}}-N_{\mathrm{start}}
},
\qquad
q_t\in[0,1].
\label{eq:relative_remaining_discharge_ratio}
\end{equation}
A value of \(q_t\) close to \(1\) indicates that a large proportion of
the discharge process remains, whereas a value close to \(0\)
indicates that the battery is approaching the EOD condition. Instead of directly regressing the continuous value \(q_t\), the proposed method formulates the problem as a five-stage classification task:
\begin{equation}
y_t
=
\begin{cases}
0,
& 0\leq q_t<0.2,
\quad \text{\emph{Recharge Required}},\\
1,
& 0.2\leq q_t<0.4,
\quad \text{\emph{Low}},\\
2,
& 0.4\leq q_t<0.6,
\quad \text{\emph{Moderate}},\\
3,
& 0.6\leq q_t<0.8,
\quad \text{\emph{Good}},\\
4,
& 0.8\leq q_t\leq1.0,
\quad \text{\emph{Normal}}.
\end{cases}
\label{eq:rds_label_definition}
\end{equation}
On a uniform time grid, \(q_t=(\tau_{\mathrm{EOD}}-\tau_t)/(\tau_{\mathrm{EOD}}-\tau_{\mathrm{start}})\), where \(\tau\) denotes elapsed physical time. The equivalent timestamp expression specifies how the ratio would be defined for irregularly sampled data. The realized cutoff and full episode duration are used offline to assign training, validation, and test targets. Neither quantity is supplied to the online classifier. The class names are display labels for time-fraction intervals. In particular, \emph{Recharge Required} denotes the lowest interval and has no independently validated operational threshold.

\subsection{Offline Preparation of the SOC Estimation Model}
\label{subsec:offline_model_preparation}

\begin{figure}[H]
\centering
\includegraphics[width=\textwidth]{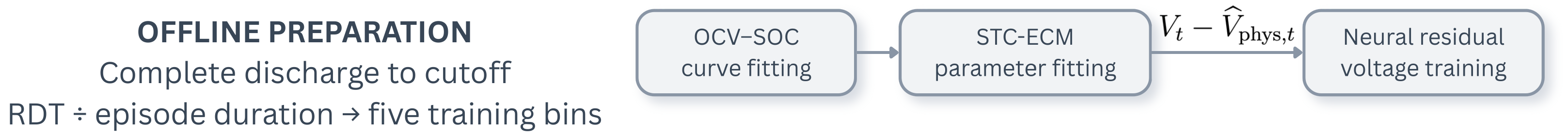}
\caption{Overall offline preparation pipeline.}
\label{fig:training_process}
\end{figure}

Before online deployment, three components are prepared using the training and validation data: (1) OCV--SOC Model Construction, (2) Offline STC-ECM Parameter Identification, and (3) Offline Residual-Voltage Model Training. The complete offline preparation pipeline is illustrated in Fig.~\ref{fig:training_process}. First, the OCV--SOC model and the STC-ECM parameters are identified with the residual-voltage correction disabled. The resulting physics-only terminal-voltage residuals, \(V_t - \widehat{V}_{\mathrm{phys},t}\), are then used as supervisory targets for training the residual-voltage model. 

\paragraph{OCV--SOC Model Construction}

The OCV--SOC model is constructed exclusively from the training data.
Because the investigated trajectories are complete discharge profiles,
the Coulomb-counting SOC is initialized at
\(S_0^{\mathrm{train}}=1\) and updated throughout each discharge
sequence. The raw samples used for OCV-model construction are expressed
as
\begin{equation}
\mathcal{D}_{\mathrm{raw}}^{\mathrm{train}}
=
\left\{
\left(
S_t^{\mathrm{train}},
V_t^{\mathrm{train}},
\left|I_t^{\mathrm{train}}\right|
\right)
\right\}_{t=1}^{N_{\mathrm{train}}},
\label{eq:ocv_raw_training_samples}
\end{equation}
where \(S_t^{\mathrm{train}}\), \(V_t^{\mathrm{train}}\), and
\(I_t^{\mathrm{train}}\) denote the Coulomb-counting SOC, measured
terminal voltage, and measured current of the \(t\)th training sample,
respectively. Since laboratory OCV measurements are unavailable, pseudo-OCV support
points are estimated from samples collected under relatively low-current
conditions. The current threshold is defined as the first quartile of
the absolute training-current distribution:
\begin{equation}
I_{\mathrm{th}}
=
Q_{0.25}
\left(
\left\{
\left|I_t^{\mathrm{train}}\right|
\right\}_{t=1}^{N_{\mathrm{train}}}
\right).
\label{eq:ocv_current_threshold}
\end{equation}
The retained low-current sample set is
\begin{equation}
\mathcal{D}_{\mathrm{low}}^{\mathrm{train}}
=
\left\{
\left(
S_t^{\mathrm{train}},
V_t^{\mathrm{train}}
\right)
:
\left|I_t^{\mathrm{train}}\right|
\leq I_{\mathrm{th}}
\right\}.
\label{eq:ocv_low_current_samples}
\end{equation}
These samples are selected because their measured terminal voltages are
less affected by ohmic and polarization voltage drops and therefore
provide closer approximations to the OCV. The SOC interval \([0,1]\) is divided into \(B=30\) uniform bins using
\(31\) bin edges:
\begin{equation}
e_k
=
\frac{k}{B},
\qquad
k=0,\ldots,B,
\label{eq:ocv_bin_edges}
\end{equation}
and the center of the \(k\)th bin is
\begin{equation}
c_k
=
\frac{e_k+e_{k+1}}{2},
\qquad
k=0,\ldots,B-1.
\label{eq:ocv_bin_centers}
\end{equation}
For each SOC bin, the SOC support point is assigned to the bin
center, while the corresponding pseudo-OCV value is calculated as the
median terminal voltage of the retained samples:
\begin{equation}
\widehat{V}_{k}^{\mathrm{OCV,train}}
=
\operatorname{median}
\left\{
V_t^{\mathrm{train}}
:
e_k
\leq
S_t^{\mathrm{train}}
<
e_{k+1},
\;
\left|I_t^{\mathrm{train}}\right|
\leq
I_{\mathrm{th}}
\right\}.
\label{eq:ocv_bin_median}
\end{equation}
The resulting SOC--OCV support set is therefore
\begin{equation}
\mathcal{D}_{\mathrm{OCV}}^{\mathrm{train}}
=
\left\{
\left(
c_k,
\widehat{V}_{k}^{\mathrm{OCV,train}}
\right)
\right\}_{k\in\mathcal{K}},
\label{eq:ocv_training_pairs}
\end{equation}
where \(\mathcal{K}\) denotes the set of SOC bins. The support pairs are sorted in ascending order of SOC. Let \(\pi(j)\)
denote the corresponding sorting permutation. The sorted SOC values are
constrained to the physical interval \([0,1]\):
\begin{equation}
\widetilde{S}_j^{\mathrm{train}}
=
\operatorname{clip}
\left(
S_{\pi(j)}^{\mathrm{train}},
0,1
\right),
\qquad
\widetilde{V}_j^{\mathrm{train}}
=
\widehat{V}_{\pi(j)}^{\mathrm{OCV,train}}.
\label{eq:ocv_training_sorting}
\end{equation}
Duplicated SOC values are removed to obtain \(M\) unique interpolation
knots:
\begin{equation}
\left\{
\left(
\overline{S}_j^{\mathrm{train}},
\overline{V}_j^{\mathrm{train}}
\right)
\right\}_{j=1}^{M},
\qquad
0
\leq
\overline{S}_1^{\mathrm{train}}
<
\cdots
<
\overline{S}_M^{\mathrm{train}}
\leq
1.
\label{eq:ocv_unique_training_knots}
\end{equation}
To enforce a nondecreasing OCV--SOC relationship, the voltage knots are
corrected using a cumulative maximum:
\begin{equation}
V_j^{\mathrm{mono}}
=
\max_{1\leq r\leq j}
\overline{V}_r^{\mathrm{train}},
\qquad
j=1,\ldots,M.
\label{eq:ocv_monotonic_correction}
\end{equation}
A shape-preserving piecewise cubic Hermite interpolating polynomial
(PCHIP)~\cite{rabbath2019comparison} is then fitted to the monotonic
knots. For
\(
S\in
[\overline{S}_j^{\mathrm{train}},
 \overline{S}_{j+1}^{\mathrm{train}}]
\),
the interval width and normalized local coordinate are defined as
\begin{equation}
h_j
=
\overline{S}_{j+1}^{\mathrm{train}}
-
\overline{S}_j^{\mathrm{train}},
\qquad
\xi
=
\frac{
S-\overline{S}_j^{\mathrm{train}}
}{
h_j
}.
\label{eq:ocv_local_coordinate}
\end{equation}
The interpolated OCV is expressed as
\begin{align}
V_{\mathrm{OC}}(S)
={}&
H_{00}(\xi)V_j^{\mathrm{mono}}
+
H_{10}(\xi)h_jm_j
\nonumber\\
&+
H_{01}(\xi)V_{j+1}^{\mathrm{mono}}
+
H_{11}(\xi)h_jm_{j+1},
\label{eq:pchip_ocv}
\end{align}
where \(m_j\) and \(m_{j+1}\) are the shape-preserving derivatives at
adjacent interpolation knots. The cubic Hermite basis functions are
\begin{equation}
\begin{aligned}
H_{00}(\xi)
&=
2\xi^3-3\xi^2+1,
\qquad
H_{10}(\xi)
=
\xi^3-2\xi^2+\xi,
\\
H_{01}(\xi)
&=
-2\xi^3+3\xi^2,
\qquad
H_{11}(\xi)
=
\xi^3-\xi^2.
\end{aligned}
\label{eq:hermite_basis}
\end{equation}
The knot derivatives are selected to preserve the local monotonicity of
the fitted OCV--SOC relationship and avoid artificial oscillations
between adjacent knots.

The final outputs of the OCV--SOC model-construction phase are the fixed
interpolation function \(V_{\mathrm{OC}}(S)\). During online estimation, it is used by the online OCV evaluation and SOC sensitivity phase.

\paragraph{Offline STC-ECM Parameter Identification}

The base electrical, hysteresis, and thermal parameters are identified offline by solving a constrained optimization problem. The parameter vector is defined as
\begin{equation}
\boldsymbol{\theta}
=
\begin{bmatrix}
R_{0,\mathrm{base}} \\
R_{1,\mathrm{base}} \\
R_{2,\mathrm{base}} \\
C_{1,\mathrm{base}} \\
C_{2,\mathrm{base}} \\
M_h \\
\gamma_h \\
C_{\mathrm{core}} \\
C_{\mathrm{surface}} \\
R_{\mathrm{core}} \\
R_{\mathrm{surface}}
\end{bmatrix},
\label{eq:physical_parameter_vector}
\end{equation}
where \(M_h\) and \(\gamma_h\) denote the hysteresis magnitude and time
constant, respectively. The parameters are identified by solving
\begin{equation}
\boldsymbol{\theta}^{\star}
=
\underset{
\boldsymbol{\theta}\in\boldsymbol{\Omega}
}{
\arg\min
}
\;
\mathcal{L}_{\mathrm{phys}}
\left(
\boldsymbol{\theta};
\mathcal{D}_{\mathrm{train}},
V_{\mathrm{OC}}
\right),
\label{eq:physical_parameter_optimization}
\end{equation}
where \(\boldsymbol{\Omega}\) denotes the physically admissible
parameter space. The physical objective evaluates the STC-ECM response
over the selected training sequences and includes an SOC-related loss
term with a weighting coefficient of \(0.1\). The bounded problem is solved using the limited-memory
Broyden--Fletcher--Goldfarb--Shanno algorithm with box constraints
(L-BFGS-B)~\cite{zhu1997algorithm}. The resistance, capacitance, and
thermal parameters are initialized at the midpoints of their
corresponding search intervals. The hysteresis magnitude and time
constant are initialized to \(0.01~\mathrm{V}\) and \(30~\mathrm{s}\),
respectively:
\begin{equation}
\boldsymbol{\theta}_0
=
\begin{bmatrix}
\mu(R_0) \\
\mu(R_1) \\
\mu(R_2) \\
\mu(C_1) \\
\mu(C_2) \\
0.01 \\
30 \\
\mu(C_{\mathrm{core}}) \\
\mu(C_{\mathrm{surface}}) \\
\mu(R_{\mathrm{core}}) \\
\mu(R_{\mathrm{surface}})
\end{bmatrix},
\label{eq:physical_initial_guess}
\end{equation}
where
\begin{equation}
\mu([a,b])
=
\frac{a+b}{2}.
\label{eq:bound_midpoint}
\end{equation}

The final output of the offline STC-ECM parameter-identification phase is the optimized parameter vector
\(\boldsymbol{\theta}^{\star}\). During online estimation, its parameters are used as the base values for the online STC-ECM parameter adaptation.

\paragraph{Offline Residual-Voltage Model Training}

After the OCV--SOC model construction and the STC-ECM parameter identification, the physics-based estimator is executed on the training trajectories with the residual-voltage correction disabled, i.e., \(\widehat{V}_{\mathrm{res},t}=0\). The corresponding physics-only terminal voltage is calculated as
\begin{align}
\widehat{V}_{\mathrm{phys},t}
={}&
\widehat{V}_{\mathrm{OC},t}
+
\Delta V_{\mathrm{OC},t}
+
h_t
\nonumber\\
&-
R_{0,t}I_t
-
\widehat{u}_{1,t|t-1}
-
\widehat{u}_{2,t|t-1}.
\label{eq:physics_only_voltage}
\end{align}
The target residual voltage is defined as the difference between the measured terminal voltage and the physics-only prediction:
\begin{equation}
V_{\mathrm{res},t}^{\mathrm{target}}
=
V_t
-
\widehat{V}_{\mathrm{phys},t}.
\label{eq:residual_voltage_target}
\end{equation}
At each sampling instant, the residual-voltage model receives the
feature vector
\begin{equation}
\mathbf{f}_t
=
\begin{bmatrix}
\widehat{S}_{t|t-1} &
\widehat{T}_{\mathrm{core},t} &
I_t &
\Delta I_t
\end{bmatrix}^{\mathrm{T}},
\qquad
\Delta I_t=I_t-I_{t-1},
\label{eq:offline_residual_features}
\end{equation}
and predicts the correction
\begin{equation}
\widehat{V}_{\mathrm{res},t}
=
\mathcal{R}_{\boldsymbol{\phi}}
\left(
\mathbf{f}_t
\right),
\label{eq:offline_residual_prediction}
\end{equation}
where \(\mathcal{R}_{\boldsymbol{\phi}}(\cdot)\) denotes the neural residual-voltage model with parameters \(\boldsymbol{\phi}\). The residual model is trained by minimizing the mean squared error
between the predicted and target residual voltages:
\begin{equation}
\mathcal{L}_{\mathrm{res}}
=
\frac{1}{N_{\mathrm{train}}}
\sum_{t=1}^{N_{\mathrm{train}}}
\left(
V_{\mathrm{res},t}^{\mathrm{target}}
-
\widehat{V}_{\mathrm{res},t}
\right)^2.
\label{eq:residual_training_loss}
\end{equation}
The hybrid
terminal-voltage prediction used during validation and testing is
therefore
\begin{equation}
\widehat{V}_{t|t-1}
=
\widehat{V}_{\mathrm{phys},t}
+
\widehat{V}_{\mathrm{res},t}.
\label{eq:offline_hybrid_voltage}
\end{equation}

The final output of the residual-voltage model-training phase is the pretrained neural residual-voltage model. During online estimation, this model is used in the pretrained residual-voltage correction phase.

\subsection{Online SOC Estimation}
\label{subsec:online_soc_estimator}

This subsection describes the online SOC estimation method illustrated in Fig.~\ref{fig:RDS_overall_architecture}. At each sampling instant \(t\), the measured current \(I_t\), terminal voltage \(V_t\), and surface temperature \(T_{\mathrm{s},t}\) are processed through four main phases: (1) second-order ECM state and terminal-voltage prediction, (2) hysteresis and OCV temperature correction, (3) core-temperature estimation, and (4) AEKF measurement update and state
correction. These phases are supported by three auxiliary phases: OCV evaluation and SOC sensitivity, online STC-ECM parameter adaptation,
and pretrained residual-voltage correction.

\paragraph{Online STC-ECM Parameter Adaptation}
To capture changes in battery electrical behavior with SOC and
temperature, the ECM resistance and capacitance parameters are adapted
at each sampling instant. The online STC-ECM parameter adaptation phase improves the representation of ohmic and
polarization dynamics under varying operating and thermal conditions.

The centered SOC and core-temperature deviation are defined as
\begin{align}
\widetilde{S}_{t-1}
&=
\operatorname{clip}
\left(
\widehat{S}_{t-1|t-1},
0,1
\right)
-0.5,
\label{eq:centered_soc}
\\
\Delta T_{t-1}
&=
\widehat{T}_{\mathrm{core},t-1}
-
T_{\mathrm{ref}},
\label{eq:temperature_deviation}
\end{align}
where \(T_{\mathrm{ref}}=25\,^{\circ}\mathrm{C}\) is the reference
temperature. The shared resistance and capacitance scaling factors are defined as
\begin{align}
\lambda_{R,t}
&=
\exp
\left(
\beta_R\widetilde{S}_{t-1}
+
\alpha_R\Delta T_{t-1}
\right),
\label{eq:resistance_scaling}
\\
\lambda_{C,t}
&=
\exp
\left(
\beta_C\widetilde{S}_{t-1}
+
\alpha_C\Delta T_{t-1}
\right).
\label{eq:capacitance_scaling}
\end{align}
The resistance parameters are adapted as
\begin{equation}
R_{i,t}
=
\operatorname{sat}_{[R_{i,\min},R_{i,\max}]}
\left(
R_{i,\mathrm{base}}\lambda_{R,t}
\right),
\qquad
i\in\{0,1,2\},
\label{eq:resistance_adaptation}
\end{equation}
whereas the capacitance parameters are adapted as
\begin{equation}
C_{i,t}
=
\operatorname{sat}_{[C_{i,\min},C_{i,\max}]}
\left(
\frac{
C_{i,\mathrm{base}}
}{
\max(\lambda_{C,t},\varepsilon)
}
\right),
\qquad
i\in\{1,2\},
\label{eq:capacitance_adaptation}
\end{equation}
where \(\varepsilon=10^{-6}\) prevents numerical instability and \(\operatorname{sat}_{[a,b]}(\cdot)\) constrains its argument to the interval \([a,b]\). The exponential formulation guarantees positive scaling factors, whereas the saturation operation prevents physically unrealistic parameter values.

The final outputs of the online STC-ECM parameter-adaptation phase are the time-varying electrical parameters
\(
R_{0,t},
R_{1,t},
R_{2,t},
C_{1,t},
C_{2,t}
\).
These adapted parameters are subsequently used in the second-order ECM state prediction and terminal voltage prediction phases. In particular, \(R_{1,t}\), \(R_{2,t}\), \(C_{1,t}\), and \(C_{2,t}\) determine the RC time constants and polarization-voltage dynamics, while \(R_{0,t}\) is used for terminal voltage prediction and internal heat generation.

\paragraph{Second-Order ECM State Prediction}
To propagate the battery state from the previous sampling instant to
the current one, the second-order ECM uses the measured current and the
adapted electrical parameters to predict the SOC and polarization
voltages before incorporating the terminal-voltage measurement.

The electrical state vector is defined as
\begin{equation}
\mathbf{z}_t
=
\begin{bmatrix}
S_t &
u_{1,t} &
u_{2,t}
\end{bmatrix}^{\mathrm{T}},
\label{eq:ecm_state_vector}
\end{equation}
where \(S_t\) denotes the SOC and \(u_{1,t}\) and \(u_{2,t}\) denote
the polarization voltages across the first and second RC branches, respectively. Assuming that positive current represents battery discharge, the prior
SOC estimate is calculated through Coulomb counting:
\begin{equation}
\widehat{S}_{t|t-1}
=
\operatorname{clip}
\left(
\widehat{S}_{t-1|t-1}
-
\frac{
\eta I_t\Delta t
}{
3600Q_{\mathrm{n}}
},
0,1
\right),
\label{eq:soc_prediction}
\end{equation}
where \(\eta\) is the Coulombic efficiency, \(Q_{\mathrm{n}}\) is the
nominal battery capacity in ampere-hours, and \(\Delta t\) is expressed
in seconds. The time constant of the \(i\)th RC branch is
\begin{equation}
\tau_{i,t}
=
R_{i,t}C_{i,t},
\qquad
i\in\{1,2\}.
\label{eq:rc_time_constant}
\end{equation}
Using trapezoidal discretization, the polarization-voltage dynamics are
expressed as
\begin{equation}
\widehat{u}_{i,t|t-1}
=
a_{i,t}\widehat{u}_{i,t-1|t-1}
+
b_{i,t}\overline{I}_t,
\qquad
i\in\{1,2\},
\label{eq:polarization_prediction}
\end{equation}
where
\begin{equation}
\begin{aligned}
a_{i,t}
&=
\frac{
1-\dfrac{\Delta t}{2\tau_{i,t}}
}{
1+\dfrac{\Delta t}{2\tau_{i,t}}
},
\qquad
b_{i,t}
=
\frac{
\dfrac{\Delta t}{C_{i,t}}
}{
1+\dfrac{\Delta t}{2\tau_{i,t}}
},
\qquad
\overline{I}_t
=
\frac{I_t+I_{t-1}}{2}.
\end{aligned}
\label{eq:polarization_coefficients}
\end{equation}
The complete prior state estimate is
\begin{equation}
\widehat{\mathbf{z}}_{t|t-1}
=
\begin{bmatrix}
\widehat{S}_{t|t-1} &
\widehat{u}_{1,t|t-1} &
\widehat{u}_{2,t|t-1}
\end{bmatrix}^{\mathrm{T}}.
\label{eq:prior_state_vector}
\end{equation}
Ignoring the derivative of the clipping operation away from the SOC boundaries, the state-transition Jacobian is approximated as
\begin{equation}
\mathbf{F}_t
=
\operatorname{diag}
\left(
1,
a_{1,t},
a_{2,t}
\right).
\label{eq:state_transition_jacobian}
\end{equation}

The final outputs of the second-order ECM state-prediction phase are the
prior state estimate \(\widehat{\mathbf{z}}_{t|t-1}\) and the
state-transition Jacobian \(\mathbf{F}_t\). In particular,
\(\widehat{S}_{t|t-1}\) is provided to the OCV evaluation and
SOC sensitivity phase and to the residual-voltage correction phase,
while \(\widehat{u}_{1,t|t-1}\) and
\(\widehat{u}_{2,t|t-1}\) are used in the core-temperature estimation
and terminal-voltage prediction phases.

\paragraph{OCV Evaluation and SOC Sensitivity}

To convert the prior SOC estimate into the corresponding OCV and construct the AEKF measurement Jacobian, the fixed OCV--SOC model is evaluated at each sampling instant. This
phase provides both the estimated OCV required for terminal-voltage
prediction and the local OCV sensitivity required to quantify how the
predicted voltage changes with SOC during the AEKF measurement update.

The prior SOC estimate is first constrained to the physical interval:
\begin{equation}
\widehat{S}_{t|t-1}^{\mathrm{clip}}
=
\operatorname{clip}
\left(
\widehat{S}_{t|t-1},
0,1
\right).
\label{eq:online_soc_clipping}
\end{equation}
The fixed OCV--SOC model then evaluates the corresponding OCV:
\begin{equation}
\widehat{V}_{\mathrm{OC},t}
=
V_{\mathrm{OC}}
\left(
\widehat{S}_{t|t-1}^{\mathrm{clip}}
\right).
\label{eq:ocv_evaluation}
\end{equation}
The SOC sensitivity required by the AEKF measurement Jacobian is
obtained from the derivative of the PCHIP interpolant:
\begin{equation}
g_{\mathrm{OCV},t}
=
\left.
\frac{
\partial V_{\mathrm{OC}}(S)
}{
\partial S
}
\right|_{
S=\widehat{S}_{t|t-1}^{\mathrm{clip}}
}.
\label{eq:ocv_derivative}
\end{equation}
For
\(
S\in
[\overline{S}_j^{\mathrm{train}},
 \overline{S}_{j+1}^{\mathrm{train}}]
\),
the derivative is calculated as
\begin{align}
\frac{
\partial V_{\mathrm{OC}}(S)
}{
\partial S
}
=
\frac{1}{h_j}
\Big[
&H_{00}'(\xi)V_j^{\mathrm{mono}}
+
H_{10}'(\xi)h_jm_j
\nonumber\\
&+
H_{01}'(\xi)V_{j+1}^{\mathrm{mono}}
+
H_{11}'(\xi)h_jm_{j+1}
\Big],
\label{eq:pchip_ocv_derivative}
\end{align}
where
\begin{equation}
\begin{aligned}
H_{00}'(\xi)
&=
6\xi^2-6\xi,
\qquad
H_{10}'(\xi)
=
3\xi^2-4\xi+1,
\\
H_{01}'(\xi)
&=
-6\xi^2+6\xi,
\qquad
H_{11}'(\xi)
=
3\xi^2-2\xi.
\end{aligned}
\label{eq:hermite_basis_derivatives}
\end{equation}
To avoid zero or negative SOC sensitivity during the AEKF update, the derivative used by the estimator is lower-bounded as
\begin{equation}
\widehat{g}_{\mathrm{OCV},t}
=
\max
\left(
g_{\mathrm{OCV},t},
10^{-6}
\right).
\label{eq:ocv_derivative_lower_bound}
\end{equation}

The final outputs of the OCV-evaluation and SOC-sensitivity phase are the estimated open-circuit voltage \(\widehat{V}_{\mathrm{OC},t}\) and the bounded SOC sensitivity \(\widehat{g}_{\mathrm{OCV},t}\). \(\widehat{V}_{\mathrm{OC},t}\) is subsequently used in the terminal-voltage prediction, while
\(\widehat{g}_{\mathrm{OCV},t}\) is provided to the AEKF measurement-update phase to construct the measurement Jacobian.

\paragraph{Core-Temperature Estimation}
The measured surface temperature does not directly represent the
battery's internal thermal state. Therefore, the core temperature is
estimated using a lumped thermal model that combines the measured
surface temperature with the heat generated by the ohmic resistance and
the two polarization branches.

Using the adapted ECM parameters, the internal heat-generation rate is
approximated as
\begin{equation}
\dot{Q}_{\mathrm{gen},t}
=
I_t^2R_{0,t}
+
\frac{
\widehat{u}_{1,t|t-1}^{\,2}
}{
R_{1,t}
}
+
\frac{
\widehat{u}_{2,t|t-1}^{\,2}
}{
R_{2,t}
}.
\label{eq:heat_generation}
\end{equation}
The core-temperature estimate is updated using a lumped thermal model:
\begin{align}
\widehat{T}_{\mathrm{core},t}
={}&
\widehat{T}_{\mathrm{core},t-1}
+
\frac{\Delta t}{C_{\mathrm{core}}}
\left[
\frac{
T_{\mathrm{s},t}
-
\widehat{T}_{\mathrm{core},t-1}
}{
R_{\mathrm{core}}
}
+
\dot{Q}_{\mathrm{gen},t}
\right],
\label{eq:core_temperature_update}
\end{align}
where \(R_{\mathrm{core}}\) is the equivalent thermal resistance
between the battery core and surface, and \(C_{\mathrm{core}}\) is the
equivalent core thermal capacitance.

The final output of the core-temperature estimation phase is the
updated core-temperature estimate
\(\widehat{T}_{\mathrm{core},t}\). This estimate is subsequently used
to calculate the OCV temperature correction and is provided, together
with the prior SOC estimate and measured current, to the pretrained
residual-voltage correction phase. It is also retained as
\(\widehat{T}_{\mathrm{core},t-1}\) for the online STC-ECM parameter adaptation
phase at the next sampling instant.

\paragraph{Hysteresis and OCV Temperature Correction}
To improve terminal-voltage modeling under dynamic current and thermal
conditions, this phase accounts for two effects that are not represented
by the basic RC polarization states. The hysteresis model captures the
history-dependent voltage response caused by changes in current
direction and magnitude, while the OCV temperature-correction term
adjusts the open-circuit voltage according to the estimated battery core
temperature.

The voltage-hysteresis state is updated as
\begin{equation}
h_t
=
\rho_hh_{t-1}
+
\left(
1-\rho_h
\right)
M_h
\tanh
\left(
\frac{I_t}{I_{\mathrm{ref}}}
\right),
\label{eq:hysteresis_update}
\end{equation}
where \(I_{\mathrm{ref}}\) is a reference current used to obtain a
dimensionless argument for the hyperbolic tangent. The hysteresis decay factor is
\begin{equation}
\rho_h
=
\exp
\left(
-\frac{\Delta t}{\gamma_h}
\right),
\label{eq:hysteresis_decay}
\end{equation}
where \(\gamma_h\) and \(M_h\) denote the hysteresis time constant and
maximum hysteresis magnitude, respectively. The hysteresis state is
constrained as
\begin{equation}
h_t
\in
\left[
-\lvert M_h\rvert,
\lvert M_h\rvert
\right].
\label{eq:hysteresis_constraint}
\end{equation}
The temperature-dependent OCV correction is calculated using the
current core-temperature estimate:
\begin{equation}
\Delta V_{\mathrm{OC},t}
=
\kappa_T
\left(
\widehat{T}_{\mathrm{core},t}
-
T_{\mathrm{ref}}
\right),
\label{eq:ocv_temperature_correction}
\end{equation}
where \(\kappa_T\) is the OCV temperature coefficient.

The final outputs of the hysteresis and OCV temperature-correction phase are the updated hysteresis voltage \(h_t\) and the temperature-dependent OCV correction \(\Delta V_{\mathrm{OC},t}\). Both quantities are subsequently included in the terminal-voltage prediction. In addition, \(h_t\) is retained as the previous hysteresis state \(h_{t-1}\) at the next sampling instant.

\paragraph{Pretrained Residual-Voltage Correction}
The neural residual-voltage model is pretrained offline and subsequently used during online inference to compensate for unmodeled
nonlinearities and residual discrepancies in the physics-based ECM prediction. Its input feature vector is defined as
\begin{equation}
\mathbf{f}_t
=
\begin{bmatrix}
\widehat{S}_{t|t-1} &
\widehat{T}_{\mathrm{core},t} &
I_t &
\Delta I_t
\end{bmatrix}^{\mathrm{T}},
\qquad
\Delta I_t
=
I_t-I_{t-1}.
\label{eq:residual_input}
\end{equation}
The hidden representation is calculated as
\begin{equation}
\mathbf{r}_t
=
\tanh
\left(
\mathbf{W}_{\mathrm{r},1}\mathbf{f}_t
+
\mathbf{b}_{\mathrm{r},1}
\right),
\label{eq:residual_hidden}
\end{equation}
and the residual-voltage correction is
\begin{equation}
\widehat{V}_{\mathrm{res},t}
=
V_{\mathrm{res}}^{\max}
\tanh
\left(
\mathbf{W}_{\mathrm{r},2}\mathbf{r}_t
+
b_{\mathrm{r},2}
\right),
\label{eq:residual_output}
\end{equation}
where the hidden dimension was set to 16 based on a grid search over
\(\{4,8,16\}\). Hidden dimensions of 4 and 8 resulted in underfitting,
whereas 16 provided a better balance between prediction performance and
model complexity. The parameter \(V_{\mathrm{res}}^{\max}\) denotes the
residual-voltage limit.

The final output of the pretrained residual-voltage correction phase is \(\widehat{V}_{\mathrm{res},t}\). This correction is subsequently provided to the terminal-voltage prediction phase. 

\paragraph{Terminal-Voltage Prediction}

After all physical and neural correction terms are available, the
predicted terminal voltage is calculated as
\begin{align}
\widehat{V}_{t|t-1}
={}&
\widehat{V}_{\mathrm{OC},t}
+
\Delta V_{\mathrm{OC},t}
+
h_t
+
\widehat{V}_{\mathrm{res},t}
\nonumber\\
&-
R_{0,t}I_t
-
\widehat{u}_{1,t|t-1}
-
\widehat{u}_{2,t|t-1}.
\label{eq:predicted_terminal_voltage}
\end{align}

The final output of the terminal-voltage-prediction phase is the prior terminal-voltage estimate \(\widehat{V}_{t|t-1}\). This estimate is
subsequently compared with the measured terminal voltage \(V_t\) to calculate the voltage innovation \(\nu_t=V_t-\widehat{V}_{t|t-1}\), which is provided to the AEKF measurement update and state correction phase.

\paragraph{AEKF Measurement Update and State Correction}

To correct the prior SOC and polarization-voltage estimates, the AEKF
uses the terminal-voltage innovation and updates the associated
uncertainties for the next sampling instant.

The prior state covariance is predicted as
\begin{equation}
\mathbf{P}_{t|t-1}
=
\mathbf{F}_t
\mathbf{P}_{t-1|t-1}
\mathbf{F}_t^{\mathrm{T}}
+
\mathbf{Q}_t,
\label{eq:prior_covariance}
\end{equation}
where \(\mathbf{P}_{t|t-1}\) denotes the prior state covariance and \(\mathbf{Q}_t\) is the process-noise covariance matrix. The measurement Jacobian is defined as
\begin{equation}
\mathbf{H}_t
=
\begin{bmatrix}
\widehat{g}_{\mathrm{OCV},t}
&
-1
&
-1
\end{bmatrix}.
\label{eq:measurement_jacobian}
\end{equation}
The voltage innovation is
\begin{equation}
\nu_t
=
V_t
-
\widehat{V}_{t|t-1}.
\label{eq:voltage_innovation}
\end{equation}
The innovation covariance and Kalman gain are calculated as
\begin{align}
\Sigma_{\nu,t}
&=
\mathbf{H}_t
\mathbf{P}_{t|t-1}
\mathbf{H}_t^{\mathrm{T}}
+
\sigma_{V,t}^{2},
\label{eq:innovation_covariance}
\\
\mathbf{K}_t
&=
\mathbf{P}_{t|t-1}
\mathbf{H}_t^{\mathrm{T}}
\Sigma_{\nu,t}^{-1},
\label{eq:kalman_gain}
\end{align}
where \(\sigma_{V,t}^{2}\) is the adaptively estimated
terminal-voltage measurement-noise variance. The posterior state is updated as
\begin{equation}
\widehat{\mathbf{z}}_{t|t}
=
\widehat{\mathbf{z}}_{t|t-1}
+
\mathbf{K}_t\nu_t.
\label{eq:posterior_state_update}
\end{equation}
For improved numerical stability, the posterior covariance is updated
using the Joseph form:
\begin{align}
\mathbf{P}_{t|t}
={}&
\left(
\mathbf{I}_3
-
\mathbf{K}_t\mathbf{H}_t
\right)
\mathbf{P}_{t|t-1}
\left(
\mathbf{I}_3
-
\mathbf{K}_t\mathbf{H}_t
\right)^{\mathrm{T}}
\nonumber\\
&+
\mathbf{K}_t
\sigma_{V,t}^{2}
\mathbf{K}_t^{\mathrm{T}},
\label{eq:posterior_covariance}
\end{align}
where \(\mathbf{I}_3\) is the \(3\times3\) identity matrix. The process- and measurement-noise statistics are adaptively updated from the voltage innovation using predefined forgetting factors. The corrected state vector is
\begin{equation}
\widehat{\mathbf{z}}_{t|t}
=
\begin{bmatrix}
\widehat{S}_{t|t} &
\widehat{u}_{1,t|t} &
\widehat{u}_{2,t|t}
\end{bmatrix}^{\mathrm{T}}.
\label{eq:corrected_state_vector}
\end{equation}
The final posterior SOC estimate is extracted as
\begin{equation}
\widehat{S}_t
=
\widehat{S}_{t|t}
=
\begin{bmatrix}
1 & 0 & 0
\end{bmatrix}
\widehat{\mathbf{z}}_{t|t}.
\label{eq:estimated_soc_output}
\end{equation}

The final outputs of the AEKF measurement-update and state-correction phase are the posterior state estimate \(\widehat{\mathbf{z}}_{t|t}\), the posterior covariance \(\mathbf{P}_{t|t}\), and the posterior SOC estimate \(\widehat{S}_t\). The estimated SOC is combined with the measured
current, terminal voltage, and surface temperature to construct the physics-informed observation window for RDS classification. 

\subsection{Online RDS Classification}
\label{subsec:online_rds_classifier}

The posterior SOC estimate is combined with the measured signals to
form the physics-informed observation vector:
\begin{equation}
\mathbf{x}_t
=
\begin{bmatrix}
I_t &
V_t &
T_{\mathrm{s},t} &
\widehat{S}_t
\end{bmatrix}^{\mathrm{T}}.
\label{eq:observation_vector}
\end{equation}
The observation window of length \(L\) is then constructed as
\begin{equation}
\mathbf{X}_t
=
\begin{bmatrix}
\mathbf{x}_{t-L+1}^{\mathrm{T}}
\\
\mathbf{x}_{t-L+2}^{\mathrm{T}}
\\
\vdots
\\
\mathbf{x}_{t}^{\mathrm{T}}
\end{bmatrix}
\in
\mathbb{R}^{L\times4}.
\label{eq:physics_informed_window}
\end{equation}
The resulting physics-informed sequence is provided to the RDS classification model to estimate the battery's relative discharge phase.

During online inference, the RDS classification model, implemented as a
pretrained lightweight TCN, receives the most recent physics-informed
observation window
\(\mathbf{X}_t\in\mathbb{R}^{L\times4}\) and predicts the current RDS
class. The TCN extracts temporal
patterns from the measured current, terminal-voltage,
surface-temperature, and estimated SOC sequences. For a batch of size
\(B\), the input tensor is represented as
\begin{equation}
\mathbf{X}
\in
\mathbb{R}^{B\times L\times D},
\qquad
D=4,
\end{equation}
where \(L\) is the observation-window length and \(D\) is the number of
input features. Before temporal convolution, the input is transposed as
\begin{equation}
\widetilde{\mathbf{X}}
\in
\mathbb{R}^{B\times D\times L}.
\end{equation}

Let
\(\mathbf{H}^{(0)}=\widetilde{\mathbf{X}}\). Each temporal layer
consists of a one-dimensional convolution (Conv1D)~\cite{kiranyaz20191}, a rectified linear
unit (ReLU) activation~\cite{agarap2018deep}, and Dropout regularization~\cite{wager2013dropout}. The
output of the \(\ell\)th temporal convolutional layer is
\begin{equation}
\mathbf{H}^{(\ell)}
=
\operatorname{Dropout}
\left(
\operatorname{ReLU}
\left(
\operatorname{Conv1D}_{\ell}
\left(
\mathbf{H}^{(\ell-1)}
\right)
\right)
\right),
\qquad
\ell=1,\ldots,M.
\label{eq:tcn_layer}
\end{equation}
The first convolutional layer maps the \(D\) input channels to \(H\)
hidden channels, whereas the remaining layers preserve the hidden
dimension. In this study, \(M=2\), \(H=64\), the kernel size is \(3\),
the dilation factor is \(1\), and the dropout probability is \(0.3\).
Symmetric padding is applied to preserve the temporal sequence length.

After the final convolutional layer, the representation at the final
time step is selected:
\begin{equation}
\mathbf{h}_t
=
\mathbf{H}^{(M)}_{:,:,-1}
\in
\mathbb{R}^{B\times H}.
\label{eq:tcn_last_representation}
\end{equation}
For stride-one convolutions with kernel size \(k=3\) and dilations \(d_1=d_2=1\), the interior receptive field spans \(1+(k-1)(d_1+d_2)=5\) positions, from \(t-2\) to \(t+2\). At the right edge, \(t+1\) and \(t+2\) fall outside the observed window and contribute only through padding. Accordingly, the final-time readout can directly depend on \(\mathbf{x}_{t-2},\mathbf{x}_{t-1},\mathbf{x}_{t}\). The other 197 positions in the \(L=200\) window have no direct convolutional path to the prediction. Earlier measurements can influence the classifier through the recursive SOC estimates in these three observations when the estimator is propagated across the episode. Thus \(L\) is the supplied window length and governs the windowing protocol, while the deployed classifier has a three-sample direct history. The interval from the earliest directly used observation to the present is \(0.2~\mathrm{s}\) at \(10~\mathrm{Hz}\) and \(20~\mathrm{s}\) at \(0.1~\mathrm{Hz}\).
The selected representation is processed by a classification head
comprising layer normalization (LayerNorm)
\cite{xu2019understanding}, a fully connected layer, a Gaussian error linear unit (GELU) activation~\cite{hendrycks2016gaussian}, Dropout, and an
output layer:
\begin{equation}
\mathbf{o}_t
=
\mathbf{W}_2
\operatorname{Dropout}
\left[
\operatorname{GELU}
\left(
\mathbf{W}_1
\operatorname{LayerNorm}
\left(
\mathbf{h}_t
\right)
+
\mathbf{b}_1
\right)
\right]
+
\mathbf{b}_2,
\label{eq:tcn_classification_head}
\end{equation}
where
\(\mathbf{o}_t\in\mathbb{R}^{B\times K}\) denotes the classification
logits and \(K=5\). The softmax score for class \(c\) is obtained
using the softmax function:
\begin{equation}
p_t^{(c)}
=
\frac{
\exp
\left(
o_t^{(c)}
\right)
}{
\displaystyle
\sum_{j=0}^{K-1}
\exp
\left(
o_t^{(j)}
\right)
},
\qquad
c\in\{0,1,\ldots,K-1\}.
\label{eq:rds_softmax}
\end{equation}
The predicted RDS class at sampling instant \(t\) is determined as
\begin{equation}
\widehat{y}_t
=
\underset{c\in\{0,\ldots,K-1\}}{\arg\max}
\;
p_t^{(c)},
\qquad
K=5.
\label{eq:rds_classifier_prediction}
\end{equation}

Before online deployment, the measured current, terminal-voltage, and surface-temperature sequences from the training and validation sets are processed offline using the online SOC-estimation pipeline employed. The resulting posterior SOC estimates are combined with the measured signals to construct physics-informed
observation windows. The training windows are used to optimize the parameters of the lightweight TCN, whereas the validation windows are used for model selection. The proposed classification model's parameters are optimized by
minimizing the categorical cross-entropy loss between the predicted class probabilities and the ground-truth RDS labels:
\begin{equation}
\mathcal{L}_{\mathrm{RDS}}
=
-
\frac{1}{B}
\sum_{i=1}^{B}
\sum_{c=0}^{K-1}
\mathbb{I}\left(y_i=c\right)
\log p_i^{(c)},
\label{eq:rds_cross_entropy}
\end{equation}
where \(B\) is the batch size, \(K=5\) is the number of RDS classes,
\(y_i\) is the ground-truth label of the \(i\)th training sample, and
\(p_i^{(c)}\) is the predicted probability that this sample belongs to
class \(c\). The indicator function
\(\mathbb{I}(y_i=c)\) equals one when \(y_i=c\) and zero otherwise. The following section presents the experimental setup used to validate our proposed RDS classification framework.

\section{Experimental Setup}\label{Experimental_Setup}

\subsection{Datasets}
\paragraph{First Dataset}
The first dataset is the public Lithium-Ion Battery Drive Cycle Dataset
introduced by \cite{yao2025multi}. It was collected using LG INR21700
M50LT NMC cells with a nominal capacity of \(4.93~\mathrm{Ah}\), a
nominal energy of \(18.2~\mathrm{Wh}\), and a nominal voltage of
\(3.69~\mathrm{V}\). The dataset contains current, terminal-voltage,
surface-temperature, and Coulomb-counting SOC for 12 distinct
drive cycles at \(5\), \(15\), \(25\), \(35\), and
\(45\,^{\circ}\mathrm{C}\), uniformly resampled to \(10~\mathrm{Hz}\).
BCDC, LA92, CSHVC, HWFET, IM, US06, PDTCB, and OCTBC were used for
training; HHDDT and FTP-72 for validation; and FTP-75 and PDMHC for
testing. The training set covers diverse driving patterns, while the
validation and test sets each include light- and heavy-duty profiles to
evaluate generalization across different load dynamics.

\paragraph{Second Dataset}
The second dataset is the public commercial 18650-format
lithium-ion battery dataset introduced by
\cite{barkholtz2017database}. It contains measured current, terminal-voltage, surface-temperature, capacity, energy, internal resistance, operating mode, battery identity, and SOC labels in the study data at \(5\), \(15\), \(25\), \(35\), and \(45\,^{\circ}\mathrm{C}\). The training set includes LCO\_1, LCO\_2, and LCO\_3, while LCO\_4 is used for validation. LCO\_5 and LCO\_6 are reserved for testing. Each battery file contains measurements collected at all five temperatures. This cell-independent split enables the model to learn from multiple LCO cells, supports model selection using an unseen cell, and evaluates its generalization to separate test cells under different thermal conditions.

Table~\ref{tab:dataset_summary} summarizes the main physical characteristics of the two datasets. In this study, only complete
discharge profiles are considered, where each profile begins from a fully charged condition and continues until the terminal voltage reaches
the cutoff threshold. The first dataset contains only complete drive-cycle discharge profiles. In contrast, the second dataset contains both complete and partial discharge profiles. Therefore, only the complete discharge profiles from the second dataset were retained, while all partial profiles were excluded from the experiments. After removing the partial discharge profiles, the retained data from the second dataset cover the temperature range of
\(5\)--\(35\,^{\circ}\mathrm{C}\).

\begin{table}[H]
\centering
\caption{Numerical summary of the datasets used in this study.}
\label{tab:dataset_summary}
\renewcommand{\arraystretch}{1.15}
\resizebox{\textwidth}{!}{%
\begin{tabular}{lll}
\toprule
\textbf{Property} &
\textbf{First Dataset} &
\textbf{Second Dataset} \\
\midrule

Cell format &
21700 &
18650 \\

Chemistries in the experiments &
NMC &
LCO \\

Nominal capacity &
\(4.93~\mathrm{Ah}\) &
\(1.1\)--\(3.0~\mathrm{Ah}\) \\

Nominal voltage &
\(3.69~\mathrm{V}\) &
\(3.3\)--\(3.6~\mathrm{V}\) \\

Number of retained ambient temperatures &
5 &
4 \\

Temperature range &
\(5\)--\(45\,^{\circ}\mathrm{C}\) &
\(5\)--\(35\,^{\circ}\mathrm{C}\) \\

Number of discharge profiles &
12 &
Up to 5 current levels \\

Maximum discharge condition &
\(30~\mathrm{W}\) &
\(6\)--\(30~\mathrm{A}\) \\

Sampling frequency &
\(10~\mathrm{Hz}\) &
\(0.1~\mathrm{Hz}\) \\

\bottomrule
\end{tabular}%
}
\end{table}

\subsection{Experimental Configuration}
The STC-ECM parameters were identified offline using the training
set of each dataset. Table~\ref{tab:physical_parameter_bounds} summarizes the parameter search bounds and initialization values used for the two datasets. The search intervals were determined through
preliminary experiments to constrain the optimization to regions, while the listed initial values were supplied to the L-BFGS-B optimizer at the beginning of parameter identification.

\begin{table}[H]
\centering
\caption{Bounds and initial values used for STC-ECM parameter
identification in the two datasets.}
\label{tab:physical_parameter_bounds}
\resizebox{\textwidth}{!}{%
\begin{tabular}{lcccccc}
\toprule
\multirow{2}{*}{\textbf{Parameter}} &
\multicolumn{3}{c}{\textbf{First Dataset}} &
\multicolumn{3}{c}{\textbf{Second Dataset}} \\
\cmidrule(lr){2-4}
\cmidrule(lr){5-7}
&
\textbf{Lower} &
\textbf{Upper} &
\textbf{Initial} &
\textbf{Lower} &
\textbf{Upper} &
\textbf{Initial} \\
\midrule

\(R_{0,\mathrm{base}}\;(\Omega)\)
& 0.01 & 0.08 & 0.045
& 0.005 & 0.05 & 0.045 \\

\(R_{1,\mathrm{base}}\;(\Omega)\)
& 0.001 & 0.15 & 0.0755
& 0.001 & 0.08 & 0.0755 \\

\(R_{2,\mathrm{base}}\;(\Omega)\)
& 0.001 & 0.15 & 0.0755
& 0.001 & 0.08 & 0.0755 \\

\(C_{1,\mathrm{base}}\;(\mathrm{F})\)
& 1000 & 10000 & 5500
& 100 & 20000 & 5500 \\

\(C_{2,\mathrm{base}}\;(\mathrm{F})\)
& 1000 & 10000 & 5500
& 1000 & 20000 & 5500 \\

\(M_h\;(\mathrm{V})\)
& 0 & 0.05 & 0.01
& 0.002 & 0.05 & 0.01 \\

\(\gamma_h\;(\mathrm{s})\)
& 1 & 500 & 30
& 0.005 & 500 & 30 \\

\(C_{\mathrm{core}}\;(\mathrm{J/K})\)
& 40 & 100 & 70
& 28 & 100 & 70 \\

\(C_{\mathrm{surface}}\;(\mathrm{J/K})\)
& 5 & 30 & 17.5
& 4 & 30 & 17.5 \\

\(R_{\mathrm{core}}\;(\mathrm{K/W})\)
& 0.3 & 4.0 & 2.15
& 1.0 & 3.0 & 2.15 \\

\(R_{\mathrm{surface}}\;(\mathrm{K/W})\)
& 1.0 & 20.0 & 10.5
& 3.0 & 20.0 & 10.5 \\

\bottomrule
\end{tabular}%
}
\end{table}
The bounded parameter-identification problem was solved using the
L-BFGS-B algorithm. The function-value and projected-gradient
tolerances were set to \(10^{-10}\) and \(10^{-8}\), respectively.
Because the optimization was implemented using SciPy, the physical
parameter fitting was executed on the CPU. 

The SOC--temperature coupling coefficients were selected through grid search and fixed during L-BFGS-B optimization. The selected values were
\(\alpha_R=0.01~^{\circ}\mathrm{C}^{-1}\),
\(\alpha_C=-0.005~^{\circ}\mathrm{C}^{-1}\),
\(\beta_R=0.15\),
\(\beta_C=0.08\), and
\(\kappa_T=0.001~\mathrm{V}/^{\circ}\mathrm{C}\).
The \(V_{\mathrm{res}}^{\max}\) was set to
\(0.003~\mathrm{V}\). The initial terminal-voltage measurement-noise
variance was \(10^{-4}~\mathrm{V}^{2}\), while the innovation
forgetting factor and process-noise smoothing factor were set to
\(0.98\) and \(0.995\), respectively.

The lightweight TCN was implemented in PyTorch and trained using the
Adam optimizer with a learning rate of \(5\times10^{-4}\). The
observation-window length was set to \(L=200\), and the model was trained
for 40 epochs with a batch size of 128. The training samples were
shuffled at the beginning of each epoch, whereas the validation and test samples were processed without shuffling. The checkpoint achieving the best validation performance was retained for final testing. All
experiments were conducted on a workstation equipped with an NVIDIA
GeForce RTX 5060 GPU, an AMD Ryzen 9 9950 processor, and
\(32~\mathrm{GB}\) of RAM.

\subsection{Baseline Methods}
\label{subsec:baseline_methods}

\paragraph{SOC Estimation Baselines}
The proposed SOC-estimation configuration uses a second-order RC
equivalent-circuit model (2RC), an adaptive extended Kalman filter
(AEKF), and backward Euler discretization~\cite{higham2007strong}. For comparison, alternative
ECM structures, state-estimation methods, and discretization schemes
were evaluated. The 2RCH model~\cite{lai2020parameter} extends the conventional 2RC structure
by incorporating voltage hysteresis, while Split Deterministic Mean Filter (SDMF)~\cite{stamoulis2015optimization} was included as an
alternative state-estimation method. The AEKF was additionally
evaluated with Tustin~\cite{zhou2024enhanced}, fourth-order Runge--Kutta (RK4)~\cite{priyadarshini2025time}, and exact
discretization methods. Tustin uses a trapezoidal bilinear
approximation, RK4 numerically integrates the continuous-time state
equations using a fourth-order scheme, and the exact method applies the
analytical discrete-time solution under the assumed input conditions.

\paragraph{Temperature-Coupling Baselines}

Table~\ref{tab:vary_temp_ecm_proposed} compares two temperature-coupling
baselines with the proposed hybrid model. The Polynomial
Temperature Model represents the thermal effect using a cubic
polynomial:
\begin{equation}
T_{\mathrm{poly}}
=
a_0+a_1T+a_2T^2+a_3T^3,
\end{equation}
where \(T\) is the measured surface temperature. The Neural
Temperature Model instead learns a nonlinear temperature
representation using a lightweight feedforward network:
\begin{equation}
T_{\mathrm{NN}}
=
\mathbf{W}_2
\operatorname{ReLU}
\left(
\mathbf{W}_1T+\mathbf{b}_1
\right)
+
\mathbf{b}_2.
\end{equation}
In both baselines, the resulting temperature representation is used to
adapt the ECM resistance and capacitance parameters and to correct the
OCV. The proposed Hybrid Model combines the PCHIP-based
OCV--SOC model, SOC--temperature-dependent RC adaptation, and the
pretrained neural residual-voltage correction.

Each model is evaluated under three update configurations. In the
RC Update configuration, only the resistance and capacitance
parameters are adapted:
\begin{equation}
\alpha_R=0.01~^{\circ}\mathrm{C}^{-1},\quad
\alpha_C=-0.005~^{\circ}\mathrm{C}^{-1},\quad
\beta_R=0.15,\quad
\beta_C=0.08,\quad
\kappa_T=0.
\end{equation}
In the OCV Update configuration, RC adaptation is disabled and
only the OCV temperature correction is applied:
\begin{equation}
\alpha_R=\alpha_C=\beta_R=\beta_C=0,
\qquad
\kappa_T=0.001~\mathrm{V}/^{\circ}\mathrm{C}.
\end{equation}
The RC + OCV Update configuration jointly applies both
mechanisms:
\begin{equation}
\alpha_R=0.01~^{\circ}\mathrm{C}^{-1},\quad
\alpha_C=-0.005~^{\circ}\mathrm{C}^{-1},\quad
\beta_R=0.15,\quad
\beta_C=0.08,\quad
\kappa_T=0.001~\mathrm{V}/^{\circ}\mathrm{C}.
\end{equation}

\paragraph{RDS Classification Comparisons}
The Transformer~\cite{vaswani2017attention}, Long Short-Term Memory network (LSTM)~\cite{graves2012long}, Gated Recurrent Unit network (GRU)~\cite{dey2017gate}, and temporal convolutional network (TCN) were evaluated under two comparison settings. The first-dataset architecture ablation used the same four inputs \(\{I,V,T_{\mathrm{s}},\widehat{S}\}\) for all four models. In the original system-level comparisons, the baselines used \(\{I,V,T_{\mathrm{s}}\}\), while the proposed TCN used all four inputs. These comparisons describe the performance of the complete configurations with their respective features. The Transformer, LSTM, and GRU used two layers, hidden dimension 128, and dropout probability 0.2. The TCN configurations used two convolutional layers, hidden dimension 64, kernel size 3, and dropout probability 0.3.

\paragraph{Comparisons needed to establish incremental value over SOC}
A direct test would compare the four-input TCN with (i) five equal-width bins applied to estimated SOC, \(\min(4,\lfloor 5\,\mathrm{clip}(\widehat S_t,0,1)\rfloor)\), (ii) multinomial logistic regression using \(\widehat S_t\) alone, (iii) the same regression with recent current added, for example the signed mean and root-mean-square current over the three directly observed samples, and (iv) multinomial logistic regression using raw elapsed time since discharge start. The elapsed-time baseline must use the start known at inference and exclude the realized duration. Learned baselines should be fitted on the training partition, tuned on the validation partition, and scored at the same held-out timestamps using the same frozen SOC estimates. The available experiments do not report these four comparisons. A matched-input comparison on the second dataset is also missing.

\subsection{Evaluation Metrics}
Terminal-voltage fit is evaluated using mean absolute error (MAE), root mean square error (RMSE), and maximum absolute error (MAX). With measured and predicted voltages in volts, let \(e^V_t=10^3(V_t-\widehat V_t)\) denote the error in millivolts. The reported metrics are
\begin{equation}
\mathrm{MAE}_V=\frac{1}{N}\sum_{t=1}^{N}|e^V_t|,
\qquad
\mathrm{RMSE}_V=\sqrt{\frac{1}{N}\sum_{t=1}^{N}(e^V_t)^2},
\qquad
\mathrm{MAX}_V=\max_t|e^V_t|.
\end{equation}
The first dataset includes Coulomb-counting SOC, a reference derived from current integration. Its accuracy depends on current calibration, initial SOC, and reference capacity. The second study dataset includes SOC labels, whose provenance and independence from estimator inputs require documentation for a reference-based comparison. Small voltage errors can coexist with biased SOC because polarization, hysteresis, temperature, and the neural voltage correction can compensate for errors in the SOC-dependent OCV term.

RDS classification is evaluated using accuracy, precision, recall, and F1-score. Let \(N\) denote the number of scored samples, \(y_i\in\{0,\ldots,K-1\}\) the ground-truth label, and \(\widehat{y}_i\) the predicted label, where \(K=5\). The classification accuracy is defined as
\begin{equation}
\mathrm{Acc.}
=
\frac{1}{N}
\sum_{i=1}^{N}
\mathbb{I}
\left(
y_i=\widehat{y}_i
\right)
\times 100\%.
\end{equation}
For each RDS class \(c\), the precision and recall are calculated as
\begin{equation}
\mathrm{Prec.}_c
=
\frac{\mathrm{TP}_c}
{\mathrm{TP}_c+\mathrm{FP}_c}\times 100\%,
\qquad
\mathrm{Rec.}_c
=
\frac{\mathrm{TP}_c}
{\mathrm{TP}_c+\mathrm{FN}_c}\times 100\%,
\end{equation}
where \(\mathrm{TP}_c\), \(\mathrm{FP}_c\), and \(\mathrm{FN}_c\)
denote the numbers of true-positive, false-positive, and false-negative
predictions for class \(c\), respectively. The class-wise F1-score is defined as
\begin{equation}
\mathrm{F1}_c
=
\frac{
2\,\mathrm{Prec.}_c\,\mathrm{Rec.}_c
}{
\mathrm{Prec.}_c+\mathrm{Rec.}_c
}.
\end{equation}

The following section reports and analyzes the experimental results of the proposed RDS classification framework.

\section{Experimental Results}\label{Experiments_Results}
The first dataset provides the dynamic drive profiles used for the component comparisons. Where indicated, its results are summarized as means and standard deviations across five ambient temperatures from \(5\) to \(45\,^{\circ}\mathrm{C}\). FTP-75 and PDMHC were also used to inspect alternative configurations in the existing experiments. Consequently, comparisons on these two profiles should be interpreted as exploratory where they informed configuration choices. A confirmatory evaluation would freeze the configuration before scoring new held-out profiles.

\subsection{Terminal-Voltage Modeling on the First Dataset}
Table~\ref{tab:32_method_ecm} compares equivalent-circuit structures, filters, and discretization methods for FTP-75 and PDMHC. With backward Euler discretization, the 2RC--AEKF configuration produced RMSEs of \(4.6\pm0.1088~\mathrm{mV}\) on FTP-75 and \(0.9\pm0.0207~\mathrm{mV}\) on PDMHC. Voltage errors serve as a measure of terminal-voltage fit. SOC error requires comparison against a documented SOC reference.

\begin{table}[H]
\centering
\caption{Comparison of terminal-voltage prediction performance for the 2RC and 2RCH
models with different filters and discretization methods on the first dataset.}
\label{tab:32_method_ecm}

\small
\renewcommand{\arraystretch}{1.15}

\resizebox{\textwidth}{!}{%
\begin{tabular}{l l l c c c c c c}
\toprule

\multirow{2}{*}{\textbf{ECM Type}} &
\multirow{2}{*}{\textbf{Filter}} &
\multirow{2}{*}{\textbf{Discretization}} &
\multicolumn{3}{c}{\textbf{FTP-75}} &
\multicolumn{3}{c}{\textbf{PDMHC}} \\

\cmidrule(lr){4-6}
\cmidrule(lr){7-9}

&
&
&
\textbf{MAE [mV]} &
\textbf{RMSE [mV]} &
\textbf{MAX [mV]} &
\textbf{MAE [mV]} &
\textbf{RMSE [mV]} &
\textbf{MAX [mV]} \\

\midrule

\multirow{8}{*}{2RC}

& \multirow{4}{*}{SDMF}
& Backward Euler
& 70.2470 \(\pm\) 1.6725
& 79.7 \(\pm\) 1.8981
& 280.4 \(\pm\) 6.6769
& 72.9404 \(\pm\) 1.7367
& 89.0 \(\pm\) 2.1190
& 295.1 \(\pm\) 7.0256 \\

& & Tustin
& 70.6370 \(\pm\) 1.6818
& 80.1 \(\pm\) 1.9064
& 290.9 \(\pm\) 6.9258
& 72.0061 \(\pm\) 1.7144
& 88.1 \(\pm\) 2.0979
& 286.8 \(\pm\) 6.8294 \\

& & RK4
& 71.2875 \(\pm\) 1.6973
& 80.7 \(\pm\) 1.9208
& 287.1 \(\pm\) 6.8362
& 75.6083 \(\pm\) 1.8002
& 92.9 \(\pm\) 2.2109
& 308.2 \(\pm\) 7.3383 \\

& & Exact
& 70.1864 \(\pm\) 1.6711
& 79.6 \(\pm\) 1.8959
& 292.3 \(\pm\) 6.9587
& 74.2855 \(\pm\) 1.7687
& 90.1 \(\pm\) 2.1460
& 298.5 \(\pm\) 7.1068 \\

\cmidrule(lr){2-9}

& \multirow{4}{*}{AEKF}

& Backward Euler (Ours)
& \textbf{0.6316 \(\pm\) 0.0150}
& \textbf{4.6 \(\pm\) 0.1088}
& \textbf{251.2 \(\pm\) 5.9798}
& \textbf{0.2032 \(\pm\) 0.0048}
& \textbf{0.9 \(\pm\) 0.0207}
& \textbf{37.3 \(\pm\) 0.8875} \\

& & Tustin
& 0.6377 \(\pm\) 0.0152
& 4.6 \(\pm\) 0.1097
& 251.9 \(\pm\) 5.9972
& 0.2047 \(\pm\) 0.0049
& 0.9 \(\pm\) 0.0209
& 37.3 \(\pm\) 0.8875 \\

& & RK4
& 0.6376 \(\pm\) 0.0152
& 4.6 \(\pm\) 0.1097
& 251.9 \(\pm\) 5.9970
& 0.2046 \(\pm\) 0.0049
& 0.9 \(\pm\) 0.0209
& 37.3 \(\pm\) 0.8875 \\

& & Exact
& 0.6361 \(\pm\) 0.0151
& 4.6 \(\pm\) 0.1096
& 252.0 \(\pm\) 5.9998
& 0.2041 \(\pm\) 0.0049
& 0.9 \(\pm\) 0.0209
& 37.3 \(\pm\) 0.8875 \\







\midrule

\multirow{8}{*}{2RCH}

& \multirow{4}{*}{SDMF}

& Backward Euler
& 160.6122 \(\pm\) 3.8241
& 198.5 \(\pm\) 4.7262
& 475.0 \(\pm\) 11.3101
& 115.0947 \(\pm\) 2.7404
& 147.0 \(\pm\) 3.5009
& 360.1 \(\pm\) 8.5735 \\

& & Tustin
& 158.7894 \(\pm\) 3.7807
& 195.7 \(\pm\) 4.6592
& 478.3 \(\pm\) 11.3872
& 117.2528 \(\pm\) 2.7917
& 149.5 \(\pm\) 3.5599
& 381.9 \(\pm\) 9.0934 \\

& & RK4
& 158.5901 \(\pm\) 3.7760
& 195.3 \(\pm\) 4.6493
& 476.1 \(\pm\) 11.3366
& 118.7165 \(\pm\) 2.8266
& 150.8 \(\pm\) 3.5895
& 375.3 \(\pm\) 8.9369 \\

& & Exact
& 159.1867 \(\pm\) 3.7902
& 196.6 \(\pm\) 4.6814
& 471.1 \(\pm\) 11.2172
& 116.8722 \(\pm\) 2.7827
& 149.4 \(\pm\) 3.5577
& 361.1 \(\pm\) 8.5968 \\

\cmidrule(lr){2-9}

& \multirow{4}{*}{AEKF}

& Backward Euler
& 2.7054 \(\pm\) 0.0644
& 11.5 \(\pm\) 0.2729
& 288.8 \(\pm\) 6.8766
& 1.5405 \(\pm\) 0.0367
& 10.3 \(\pm\) 0.2455
& 290.9 \(\pm\) 6.9272 \\

& & Tustin
& 2.7257 \(\pm\) 0.0649
& 11.5 \(\pm\) 0.2744
& 288.7 \(\pm\) 6.8748
& 1.5573 \(\pm\) 0.0371
& 10.4 \(\pm\) 0.2473
& 290.8 \(\pm\) 6.9238 \\

& & RK4
& 2.7255 \(\pm\) 0.0649
& 11.5 \(\pm\) 0.2744
& 288.7 \(\pm\) 6.8748
& 1.5571 \(\pm\) 0.0371
& 10.4 \(\pm\) 0.2473
& 290.8 \(\pm\) 6.9238 \\

& & Exact
& 2.7223 \(\pm\) 0.0648
& 11.5 \(\pm\) 0.2742
& 288.6 \(\pm\) 6.8721
& 1.5545 \(\pm\) 0.0370
& 10.4 \(\pm\) 0.2470
& 290.8 \(\pm\) 6.9228 \\







\bottomrule
\end{tabular}}
\end{table}

Table~\ref{tab:vary_temp_ecm_proposed} compares three temperature-coupling models under RC-only, OCV-only, and joint RC--OCV updates. The joint hybrid configuration produced the lowest reported MAE and RMSE for both drive profiles. Its mean voltage MAE was \(0.5189~\mathrm{mV}\) on FTP-75 and \(0.3864~\mathrm{mV}\) on PDMHC.

\begin{table}[H]
\centering
\caption{Terminal-voltage prediction performance of the polynomial
temperature model, neural temperature model, and proposed hybrid model
under RC-only, OCV-only, and joint RC--OCV temperature-update
configurations on the first dataset.}
\label{tab:vary_temp_ecm_proposed}
\small
\renewcommand{\arraystretch}{1.15}

\resizebox{\textwidth}{!}{%
\begin{tabular}{l l c c c c c c}
\toprule

\multirow{2}{*}{\textbf{Model}} &
\multirow{2}{*}{\textbf{Update Type}} &
\multicolumn{3}{c}{\textbf{FTP-75}} &
\multicolumn{3}{c}{\textbf{PDMHC}} \\

\cmidrule(lr){3-5}
\cmidrule(lr){6-8}

&
&
\textbf{MAE [mV]} &
\textbf{RMSE [mV]} &
\textbf{MAX [mV]} &
\textbf{MAE [mV]} &
\textbf{RMSE [mV]} &
\textbf{MAX [mV]} \\

\midrule

\multirow{3}{*}{Polynomial Temperature Model}
& RC Update
& \(22.5998 \pm 3.2050\)
& \(35.2476 \pm 12.3578\)
& \(453.9707 \pm 67.9803\)
& \(18.7525 \pm 0.8109\)
& \(30.1871 \pm 2.6568\)
& \(435.0426 \pm 27.7630\) \\

& OCV Update
& \(22.5587 \pm 2.7928\)
& \(35.4468 \pm 12.4080\)
& \(471.1032 \pm 62.6695\)
& \(18.4161 \pm 1.1621\)
& \(30.2339 \pm 3.1373\)
& \(443.6866 \pm 33.7626\) \\

& RC + OCV Update
& \(22.6929 \pm 2.8454\)
& \(35.4646 \pm 12.2896\)
& \(455.0236 \pm 70.8208\)
& \(18.8362 \pm 1.2595\)
& \(30.4053 \pm 2.5976\)
& \(438.8881 \pm 29.0894\) \\

\midrule

\multirow{3}{*}{Neural Temperature Model}
& RC Update
& \(25.2065 \pm 6.2347\)
& \(37.5337 \pm 14.1825\)
& \(411.0904 \pm 60.2398\)
& \(28.0822 \pm 9.9072\)
& \(39.9967 \pm 12.7434\)
& \(367.8225 \pm 35.4247\) \\

& OCV Update
& \(28.7108 \pm 10.3545\)
& \(40.4571 \pm 16.2769\)
& \(441.2209 \pm 46.2954\)
& \(24.8833 \pm 6.1156\)
& \(35.9754 \pm 7.9708\)
& \(410.1656 \pm 20.4329\) \\

& RC + OCV Update
& \(31.1028 \pm 10.0024\)
& \(43.4566 \pm 15.9576\)
& \(380.7916 \pm 60.0008\)
& \(37.4592 \pm 11.1930\)
& \(49.8778 \pm 14.1034\)
& \(327.7206 \pm 35.2477\) \\

\midrule

\multirow{3}{*}{Hybrid Model}
& RC Update
& \(0.6920 \pm 0.4180\)
& \(3.1475 \pm 1.5638\)
& \(162.0635 \pm 53.8767\)
& \(0.5574 \pm 0.2516\)
& \(3.1999 \pm 1.6188\)
& \(169.8329 \pm 39.8059\) \\

& OCV Update
& \(0.5258 \pm 0.2050\)
& \(2.9078 \pm 1.2732\)
& \(168.3609 \pm 59.8809\)
& \(0.4582 \pm 0.2542\)
& \(3.1139 \pm 1.7071\)
& \(181.2592 \pm 41.3030\) \\

& RC + OCV Update (Ours)
& \(\mathbf{0.5189 \pm 0.1702}\)
& \(\mathbf{2.8153 \pm 1.3019}\)
& \(\mathbf{160.1664 \pm 68.2927}\)
& \(\mathbf{0.3864 \pm 0.1093}\)
& \(\mathbf{2.6400 \pm 0.9642}\)
& \(\mathbf{163.1010 \pm 37.5258}\) \\

\bottomrule
\end{tabular}%
}
\end{table}

For the joint RC and OCV update, the values in Table~\ref{tab:vary_temp_ecm_proposed} are the reported five-temperature summaries. The corresponding RMSEs are \(2.8153\pm1.3019~\mathrm{mV}\) for FTP-75 and \(2.6400\pm0.9642~\mathrm{mV}\) for PDMHC. These are terminal-voltage errors. They do not establish SOC accuracy.

\subsection{Feature and Architecture Ablations on the First Dataset}

\begin{figure}[H]
\centering
\includegraphics[width=1\textwidth]{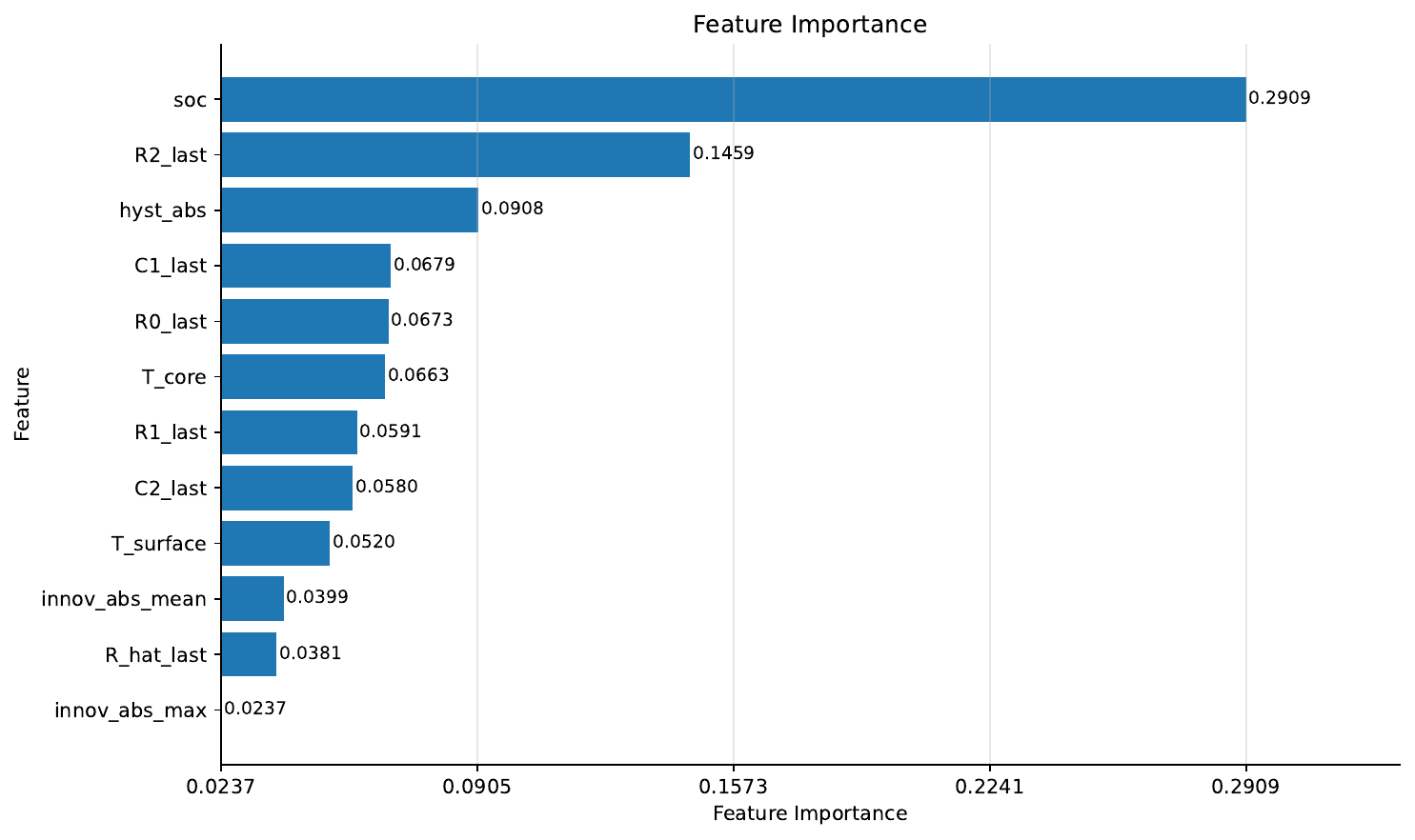}
\caption{XGBoost-based feature importance for RDS classification with $K=5$ bins.}
\label{fig:feature_importance}
\end{figure}

To assess the relevance of the internal states produced by the proposed
SOC estimation method, an XGBoost classifier was trained for the
K=5 RDS classification task using the estimated state variables as
inputs. Figure~\ref{fig:feature_importance} reports the corresponding
feature-importance scores. The estimated SOC achieves the highest
importance score of \(0.2909\), followed by \(R_{2,t}\) with \(0.1459\)
and the absolute hysteresis state with \(0.0908\). The remaining ECM,
thermal, innovation, and noise-related variables contribute smaller but
non-negligible importance scores. These results indicate that the
estimated SOC is the most informative state variable for RDS
classification, which motivates its inclusion in the proposed
physics-informed observation window. Nevertheless, feature importance
does not directly quantify the benefit of combining multiple state
variables. Therefore, the following experiment further evaluates
different combinations of estimated states as inputs to the neural RDS
classifier.

\begin{figure}[H]
\centering
\includegraphics[width=\textwidth]{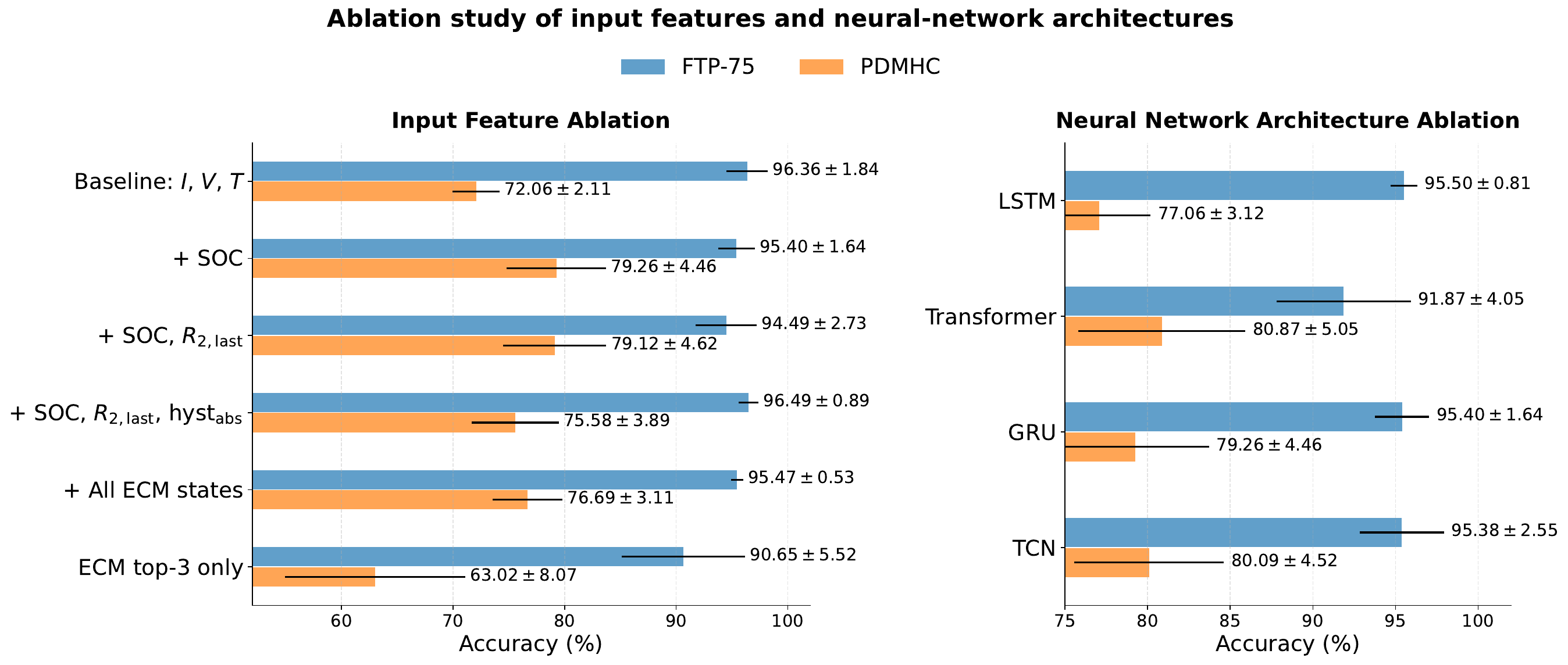}
\caption{RDS classification accuracy for the input-feature and architecture ablations.}
\label{fig:feature_model_ablation}
\end{figure}

Figure~\ref{fig:feature_model_ablation} summarizes the input-feature and architecture ablations. Within the GRU comparison, adding estimated SOC to the sensor inputs produced the highest reported average accuracy across the two profiles among the evaluated feature combinations. Because the comparison starts from a sensor-input model, it measures the value of adding SOC to those signals. The inverse question, whether the signals add value beyond SOC, requires an SOC-only baseline.

Table~\ref{tab:matched_input_first} presents the existing architecture-ablation results with the same four inputs for every model. The LSTM gave the highest observed FTP-75 accuracy. The Transformer gave the highest observed PDMHC accuracy and F1 score, while the TCN gave the highest mean accuracy across the two profiles by a small margin. Given the reported variation across temperatures and the exploratory choice of configurations, the table does not establish that any one architecture is superior. The TCN and GRU entries here come from the architecture ablation. The TCN's separately reported end-to-end results in Table~\ref{tab:final_results} are retained as a distinct run.

\begin{table}[H]
\centering
\caption{Matched-input five-stage architecture ablation on the first dataset. All models receive \(\{I,V,T_{\mathrm{s}},\widehat S\}\). Values are mean \(\pm\) standard deviation across five temperatures.}
\label{tab:matched_input_first}
\resizebox{\textwidth}{!}{%
\begin{tabular}{lcccccccc}
\toprule
\multirow{2}{*}{\textbf{Neural Network}} &
\multicolumn{4}{c}{\textbf{FTP-75}} &
\multicolumn{4}{c}{\textbf{PDMHC}} \\
\cmidrule(lr){2-5}
\cmidrule(lr){6-9}

&
\textbf{Acc. (\%)} &
\textbf{Prec. (\%)} &
\textbf{Rec. (\%)} &
\textbf{F1 (\%)} &
\textbf{Acc. (\%)} &
\textbf{Prec. (\%)} &
\textbf{Rec. (\%)} &
\textbf{F1 (\%)} \\
\midrule

LSTM        & 95.4998 \(\pm\) 0.8096 & 95.6455 \(\pm\) 0.7596 & 95.5000 \(\pm\) 0.8082 & 95.4933 \(\pm\) 0.8088 & 77.0591 \(\pm\) 3.1178 & 80.1061 \(\pm\) 4.4624 & 77.1069 \(\pm\) 3.1144 & 76.4842 \(\pm\) 2.8231 \\

Transformer & 91.8653 \(\pm\) 4.0502 & 92.2678 \(\pm\) 3.9984 & 91.8629 \(\pm\) 4.0452 & 91.8389 \(\pm\) 4.0792 & 80.8663 \(\pm\) 5.0453 & 83.8916 \(\pm\) 5.1029 & 80.9042 \(\pm\) 5.0370 & 80.3406 \(\pm\) 5.1440 \\

GRU         & 95.4035 \(\pm\) 1.6368 & 95.5210 \(\pm\) 1.6150 & 95.4049 \(\pm\) 1.6353 & 95.4009 \(\pm\) 1.6568 & 79.2567 \(\pm\) 4.4613 & 82.3872 \(\pm\) 3.8551 & 79.3017 \(\pm\) 4.4510 & 78.6258 \(\pm\) 4.7419 \\

TCN & 95.3810 \(\pm\) 2.5478 & 95.5747 \(\pm\) 2.4500 & 95.3842 \(\pm\) 2.5470 & 95.3833 \(\pm\) 2.5614 & 80.0866 \(\pm\) 4.5224 & 83.4871 \(\pm\) 3.0055 & 80.1303 \(\pm\) 4.5116 & 79.4505 \(\pm\) 4.8660 \\

\bottomrule
\end{tabular}%
}
\end{table}

\begin{figure}[H]
\centering
\includegraphics[width=1\textwidth]{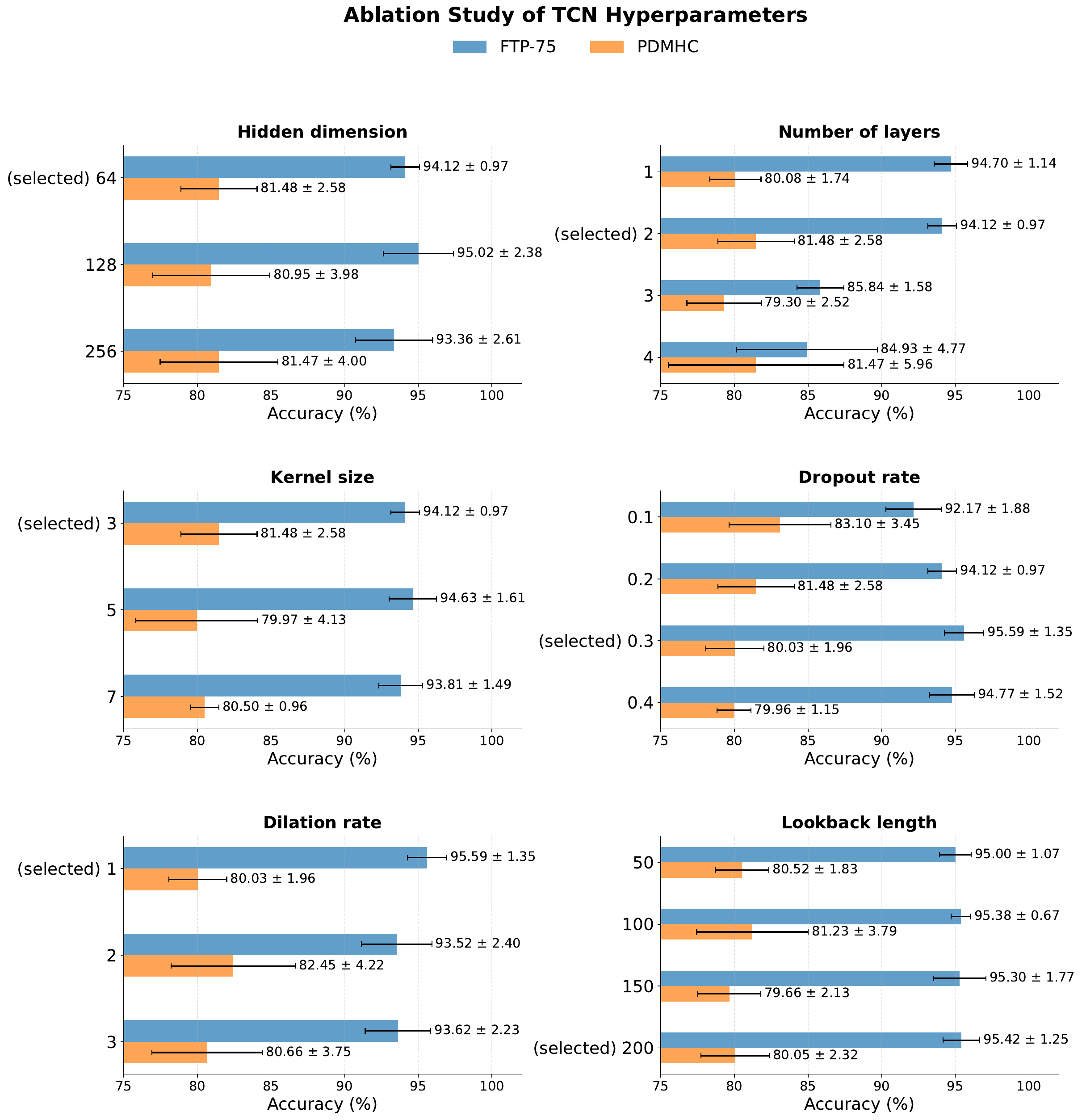}
\caption{TCN hyperparameter ablation results.}
\label{fig:tcn_hyperparameter_ablation_all_in_one}
\end{figure}

Figure~\ref{fig:tcn_hyperparameter_ablation_all_in_one} summarizes the exploratory TCN configuration comparisons. The reported end-to-end configuration uses 64 hidden channels, two convolutional layers, kernel size 3, dropout probability 0.3, dilation 1, and \(L=200\). Varying \(L\) can change eligible timestamps and the amount of data available for the experiment. For the specified two-layer readout, the receptive-field calculation establishes a direct dependency on three observed samples.

\subsection{End-to-End System Results on the First Dataset}
Table~\ref{tab:final_results} records the original end-to-end comparison. The Transformer, LSTM, GRU, and sensor-only TCN use \(\{I,V,T_{\mathrm{s}}\}\), while the proposed TCN additionally receives \(\widehat S\). The difference in inputs prevents attribution of the performance differences to the classifier architecture. For the proposed TCN, five-stage accuracies were \(95.38\%\) on FTP-75 and \(81.23\%\) on PDMHC. Classification accuracy was generally lower for ten bins, particularly on PDMHC. Changing the number of bins changes the classification task and the granularity at which errors are counted.

\begin{table}[H]
\centering
\caption{Original end-to-end system results for ten and five RDS bins. The proposed TCN receives estimated SOC in addition to the three sensor inputs used by the other models. Table~\ref{tab:matched_input_first} reports the separate matched-input architecture ablation.}
\label{tab:final_results}

\resizebox{\textwidth}{!}{%
\begin{tabular}{l l c c c c c c c c}
\toprule

\multirow{2}{*}{\textbf{RDS Setting}} &
\multirow{2}{*}{\textbf{Model}} &
\multicolumn{4}{c}{\textbf{FTP-75}} &
\multicolumn{4}{c}{\textbf{PDMHC}} \\

\cmidrule(lr){3-6}
\cmidrule(lr){7-10}

&
&
\textbf{Acc. (\%)} &
\textbf{Prec. (\%)} &
\textbf{Rec. (\%)} &
\textbf{F1 (\%)} &
\textbf{Acc. (\%)} &
\textbf{Prec. (\%)} &
\textbf{Rec. (\%)} &
\textbf{F1 (\%)} \\

\midrule

\multirow{5}{*}{\shortstack{\textbf{10 stages}\\\textnormal{(comparison)}}}

& Transformer
& \(86.5666 \pm 6.1659\)
& \(87.5745 \pm 5.0526\)
& \(86.5698 \pm 6.1562\)
& \(86.4576 \pm 6.3920\)
& \(44.9899 \pm 6.1409\)
& \(48.4544 \pm 7.7170\)
& \(45.1149 \pm 6.1287\)
& \(41.7710 \pm 6.0582\) \\

& LSTM
& \(91.2789 \pm 3.4453\)
& \(91.5088 \pm 3.4172\)
& \(91.2789 \pm 3.4453\)
& \(91.2337 \pm 3.5328\)
& \(49.7452 \pm 6.2960\)
& \(55.3585 \pm 7.6541\)
& \(49.7452 \pm 6.2960\)
& \(46.8187 \pm 5.6957\) \\

& GRU
& \(91.9797 \pm 3.6782\)
& \(92.2341 \pm 3.5290\)
& \(91.9797 \pm 3.6782\)
& \(91.9353 \pm 3.7455\)
& \(48.6459 \pm 5.9007\)
& \(54.2683 \pm 8.0366\)
& \(48.6459 \pm 5.9007\)
& \(45.8862 \pm 5.5509\) \\

& TCN
& \(84.1982 \pm 3.2090\)
& \(84.8908 \pm 3.0595\)
& \(84.2058 \pm 3.2051\)
& \(84.0901 \pm 3.2577\)
& \(48.7276 \pm 4.6442\)
& \(55.3672 \pm 7.2909\)
& \(48.8439 \pm 4.6358\)
& \(46.7069 \pm 5.3356\) \\

& \textbf{Ours}
& \(\mathbf{90.3392 \pm 2.9811}\)
& \(\mathbf{90.7641 \pm 2.9754}\)
& \(\mathbf{90.3430 \pm 2.9811}\)
& \(\mathbf{90.3082 \pm 3.0031}\)
& \(\mathbf{57.0878 \pm 2.3146}\)
& \(\mathbf{65.3274 \pm 4.2918}\)
& \(\mathbf{57.1850 \pm 2.3110}\)
& \(\mathbf{54.5395 \pm 3.0134}\) \\

\midrule

\multirow{5}{*}{\shortstack{\textbf{5 stages}\\\textbf{(proposed)}}}

& Transformer
& \(95.9583 \pm 1.9841\)
& \(96.0330 \pm 1.9128\)
& \(95.9604 \pm 1.9803\)
& \(95.9490 \pm 2.0153\)
& \(69.5056 \pm 2.6816\)
& \(73.3998 \pm 3.2627\)
& \(69.5741 \pm 2.6754\)
& \(68.0365 \pm 2.7128\) \\

& LSTM
& \(95.2785 \pm 2.2944\)
& \(95.3819 \pm 2.2306\)
& \(95.2785 \pm 2.2944\)
& \(95.2507 \pm 2.3179\)
& \(73.9314 \pm 4.3894\)
& \(77.7564 \pm 6.0516\)
& \(73.9314 \pm 4.3894\)
& \(72.9409 \pm 4.1605\) \\

& GRU
& \(96.3635 \pm 1.8433\)
& \(96.3965 \pm 1.8022\)
& \(96.3635 \pm 1.8433\)
& \(96.3514 \pm 1.8589\)
& \(72.0635 \pm 2.1058\)
& \(75.6608 \pm 3.4200\)
& \(72.0635 \pm 2.1058\)
& \(70.9371 \pm 1.8832\) \\

& TCN
& \(95.3601 \pm 1.8946\)
& \(95.4348 \pm 1.8917\)
& \(95.3651 \pm 1.8920\)
& \(95.3328 \pm 1.9349\)
& \(71.8784 \pm 2.8752\)
& \(75.8889 \pm 4.5663\)
& \(71.9417 \pm 2.8675\)
& \(70.7535 \pm 2.7876\) \\

& \textbf{Ours}
& \(\mathbf{95.3808 \pm 0.6689}\)
& \(\mathbf{95.5533 \pm 0.6784}\)
& \(\mathbf{95.3797 \pm 0.6668}\)
& \(\mathbf{95.3794 \pm 0.6781}\)
& \(\mathbf{81.2306 \pm 3.7870}\)
& \(\mathbf{84.4342 \pm 2.5519}\)
& \(\mathbf{81.2500 \pm 3.7816}\)
& \(\mathbf{80.5512 \pm 4.2040}\) \\

\bottomrule
\end{tabular}%
}

\end{table}

\begin{figure}[H]
\centering

\textbf{FTP-75}\\[0.3em]

\begin{tabular}{ccccc}
\includegraphics[width=0.19\textwidth]{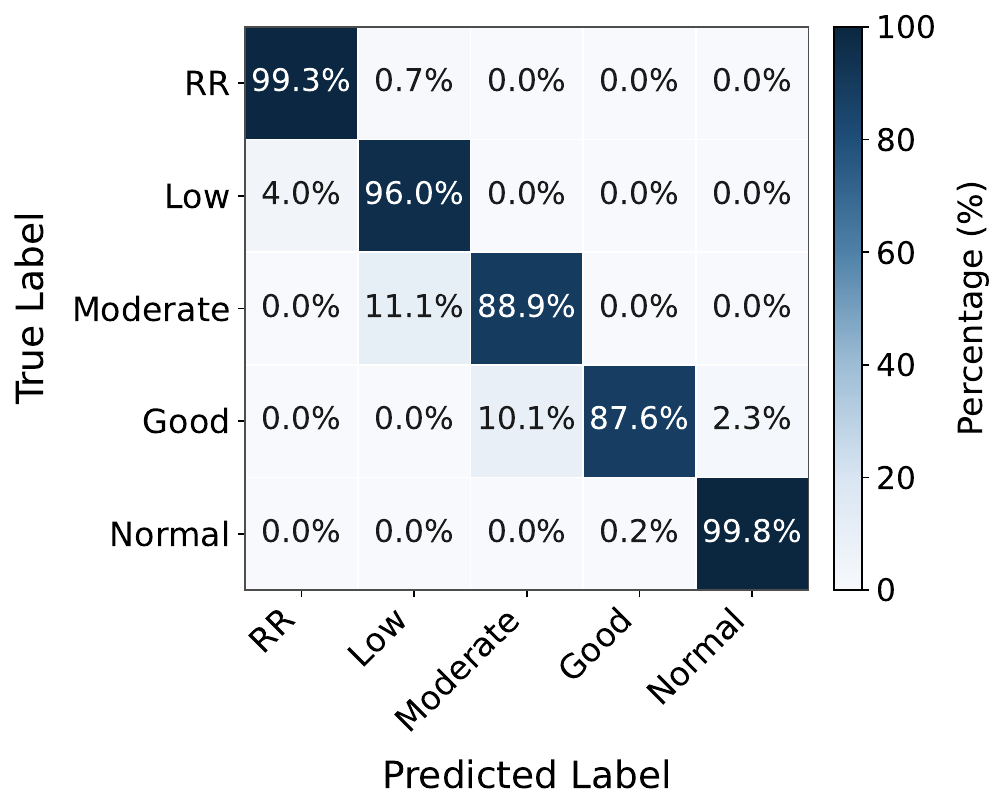} &
\includegraphics[width=0.19\textwidth]{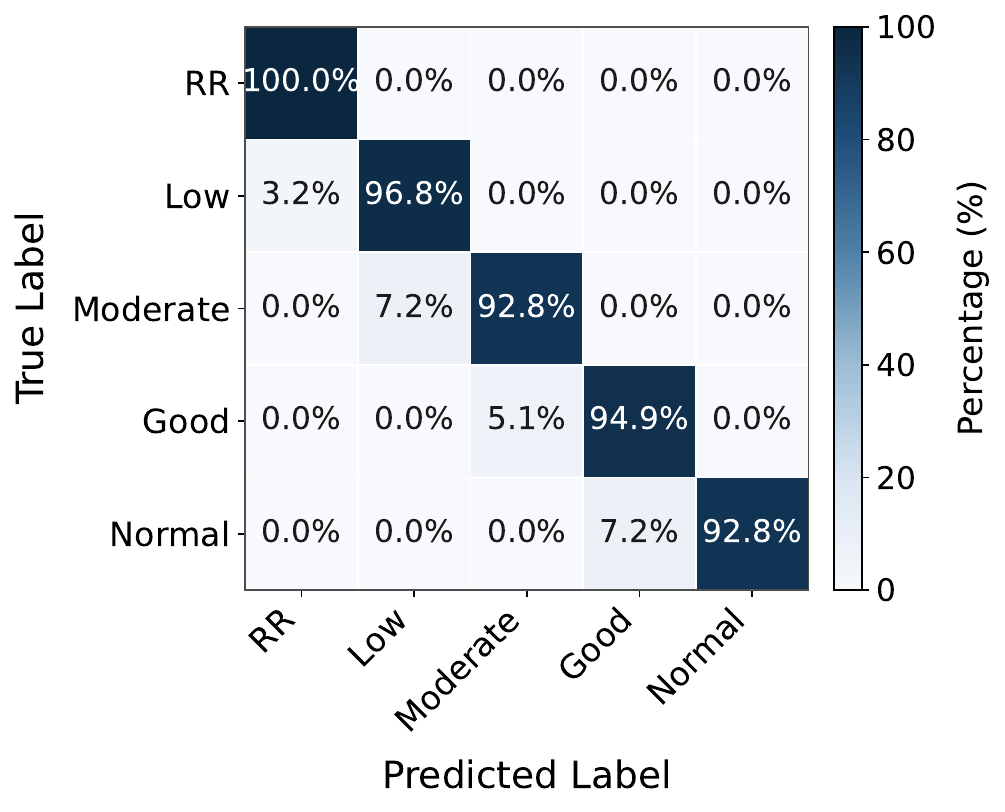} &
\includegraphics[width=0.19\textwidth]{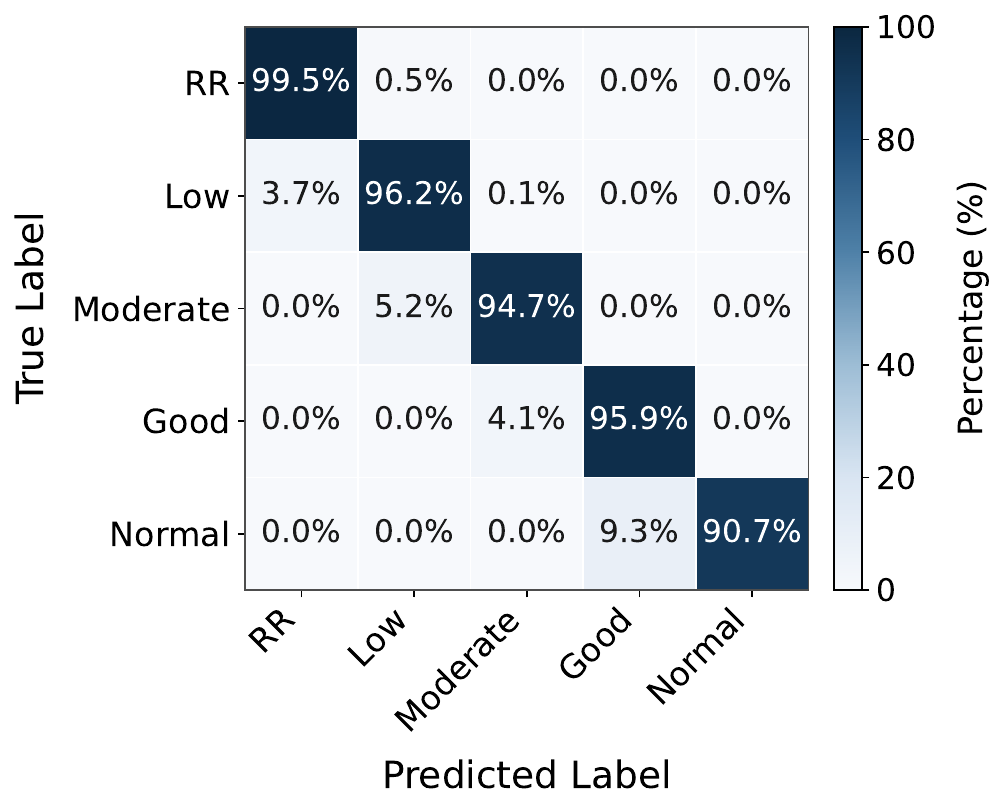} &
\includegraphics[width=0.19\textwidth]{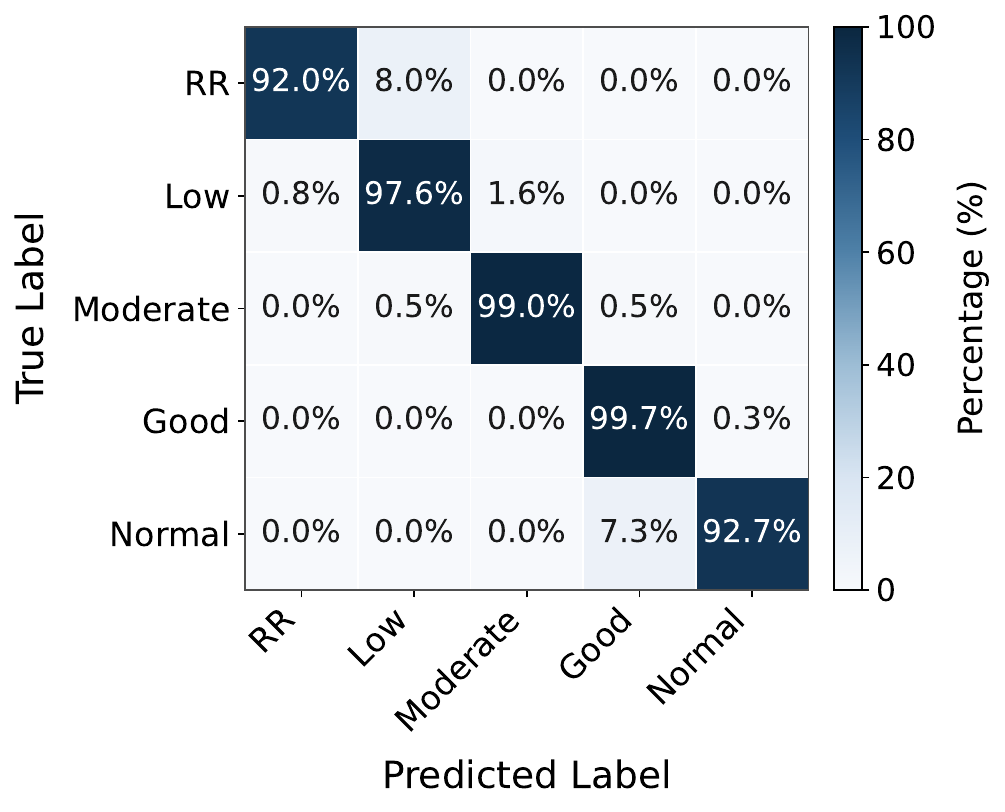} &
\includegraphics[width=0.19\textwidth]{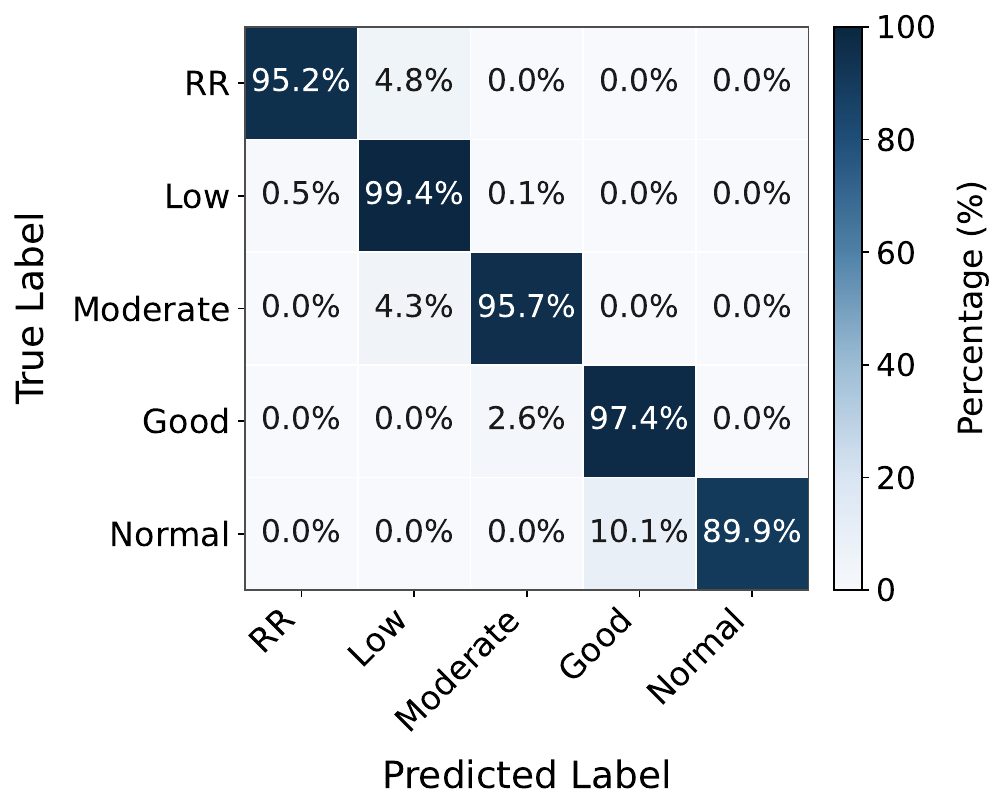}
\\[-0.2em]

{$5\,^\circ\mathrm{C}$} &
{$15\,^\circ\mathrm{C}$} &
{$25\,^\circ\mathrm{C}$} &
{$35\,^\circ\mathrm{C}$} &
{$45\,^\circ\mathrm{C}$}
\end{tabular}

\vspace{1em}

\textbf{PDMHC}\\[0.3em]

\begin{tabular}{ccccc}
\includegraphics[width=0.19\textwidth]{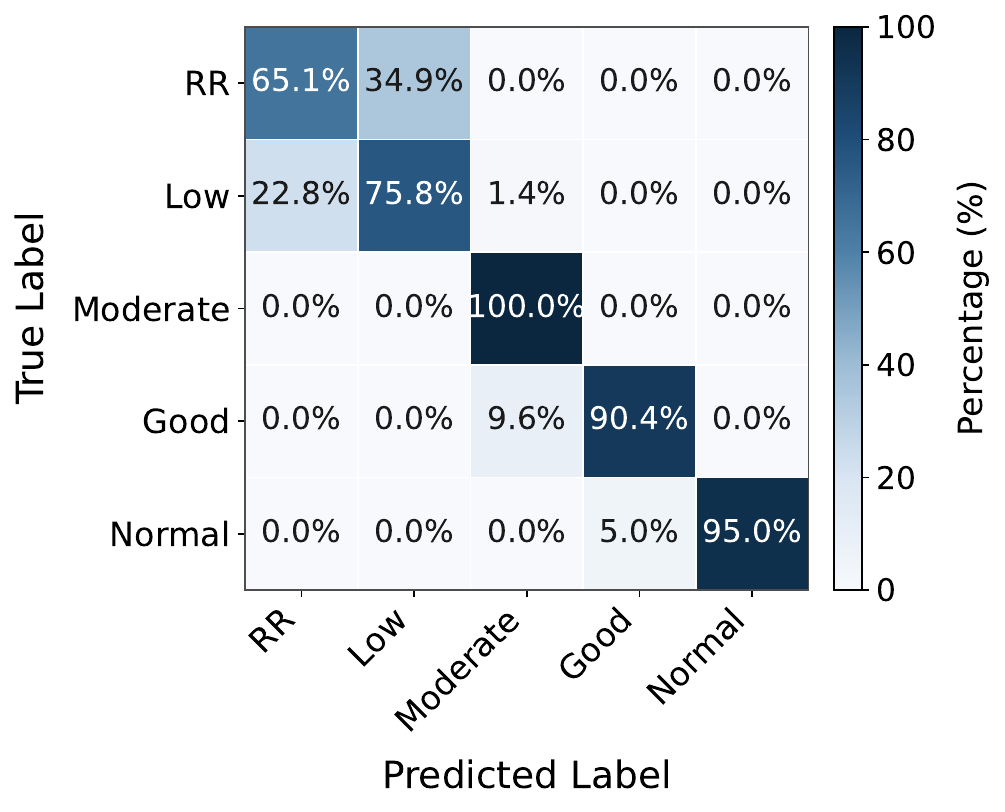} &
\includegraphics[width=0.19\textwidth]{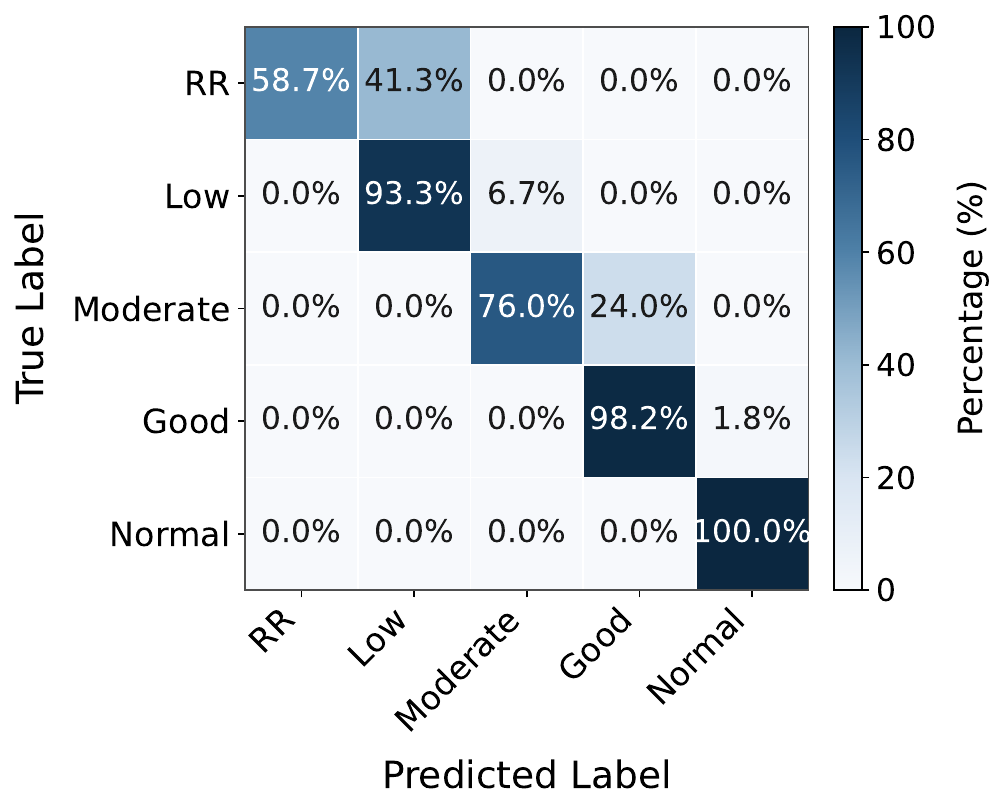} &
\includegraphics[width=0.19\textwidth]{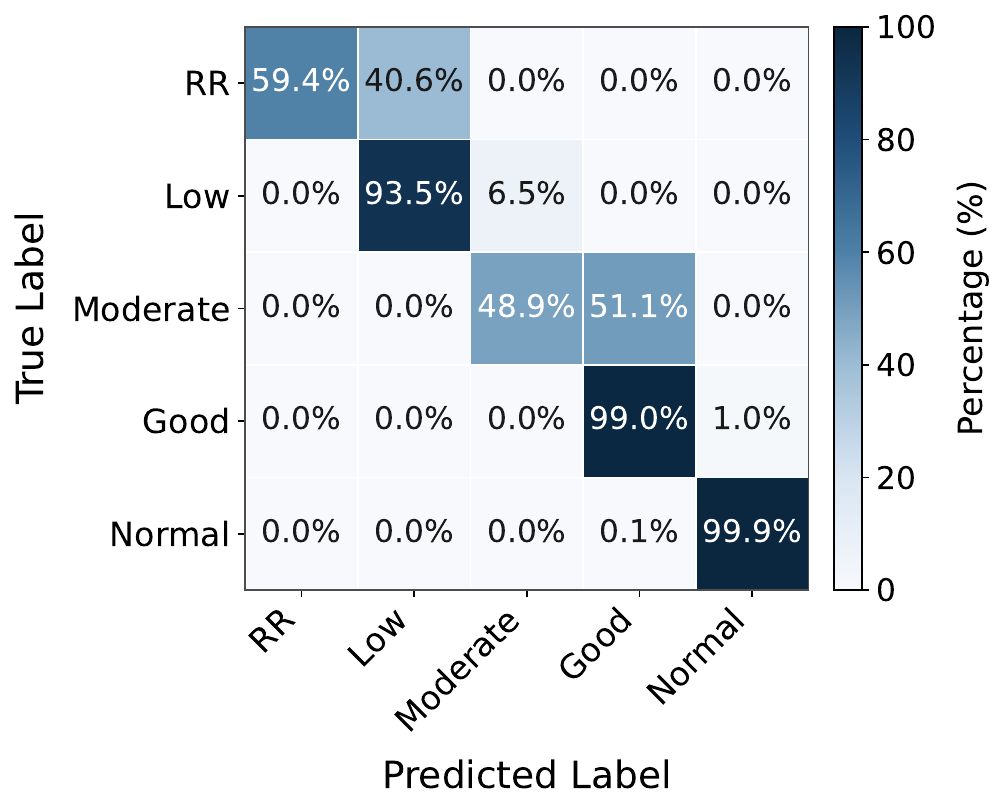} &
\includegraphics[width=0.19\textwidth]{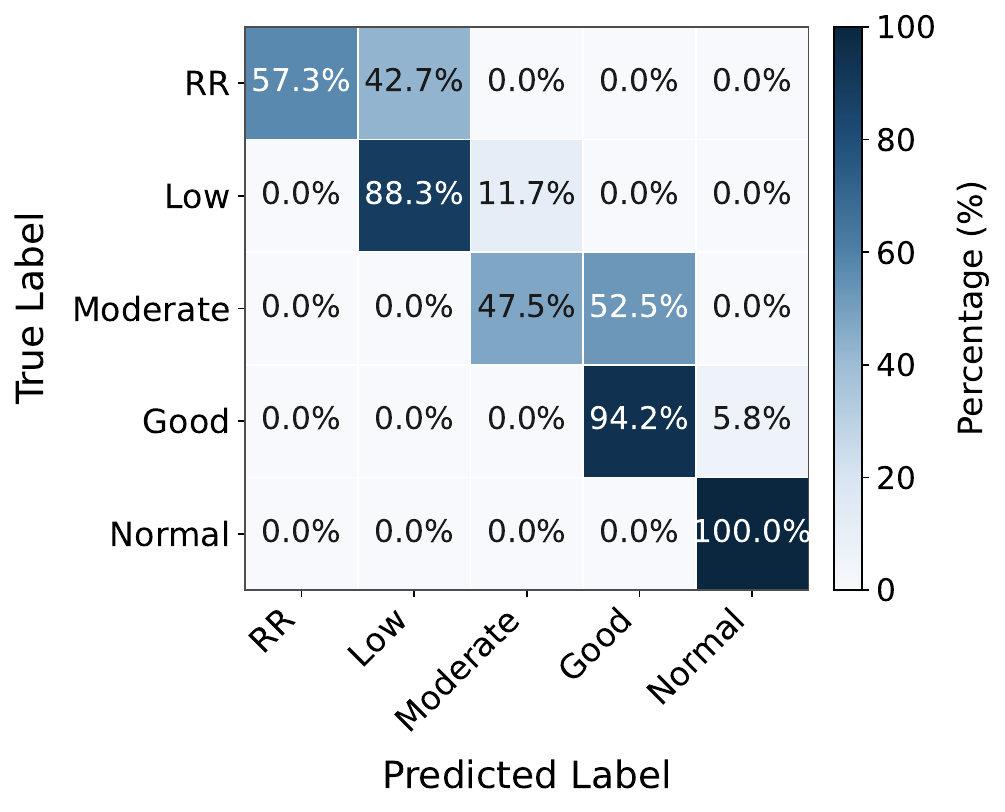} &
\includegraphics[width=0.19\textwidth]{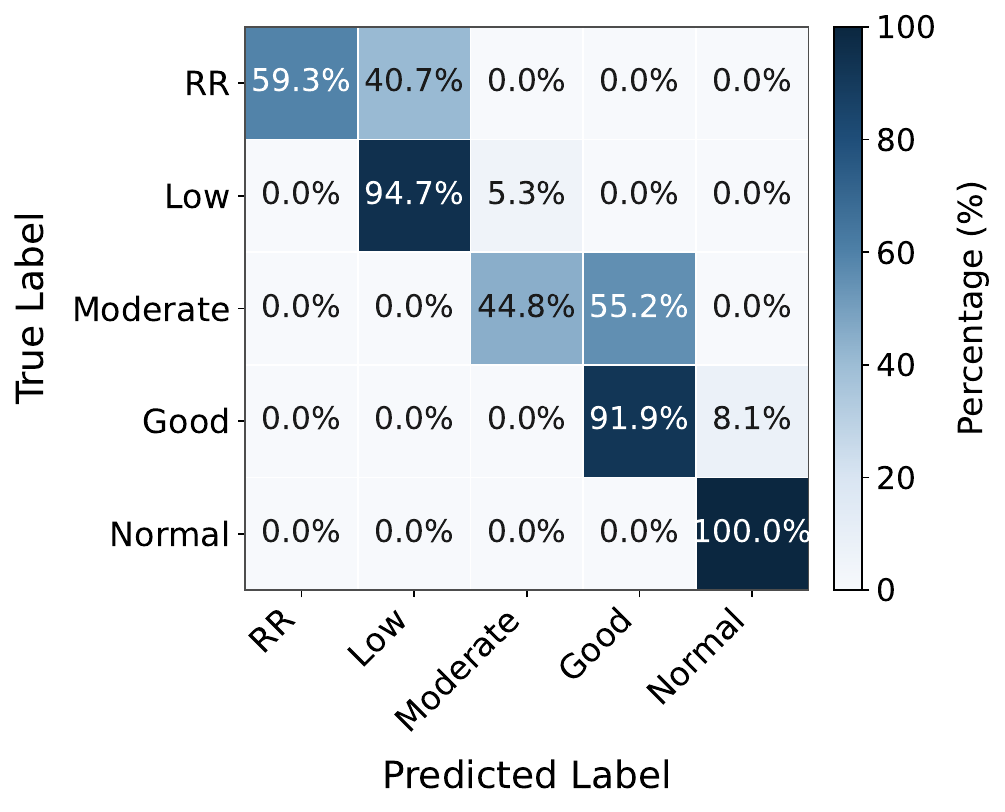}
\\[-0.2em]

{$5\,^\circ\mathrm{C}$} &
{$15\,^\circ\mathrm{C}$} &
{$25\,^\circ\mathrm{C}$} &
{$35\,^\circ\mathrm{C}$} &
{$45\,^\circ\mathrm{C}$}
\end{tabular}

\caption{Confusion matrices for RDS classification.}
\label{fig:confusion_matrices_all_bins}

\end{figure}

Figure~\ref{fig:confusion_matrices_all_bins} presents the
temperature-wise confusion matrices of the proposed five-stage RDS
classification method for the five classes: Recharge Required (RR), Low, Moderate,
Good, and Normal. For FTP-75, most predictions are concentrated along
the main diagonal at all five temperatures, indicating consistent
classification performance under varying thermal conditions. The
remaining misclassifications occur primarily between adjacent discharge
stages, particularly between Moderate and Good and between Good and
Normal.

The PDMHC profile produces greater class overlap because of its more
complex and irregular load dynamics. The most frequent errors occur
between Recharge Required and Low and between Moderate and Good,
whereas the Good and Normal stages remain comparatively well
identified. Importantly, most misclassifications occur between adjacent
RDS stages rather than distant stages. This result indicates that the
proposed classification method generally preserves the ordinal progression of the
battery discharge process, even under challenging load and temperature
conditions.

\begin{figure}[H]
\centering
\includegraphics[width=1\textwidth]{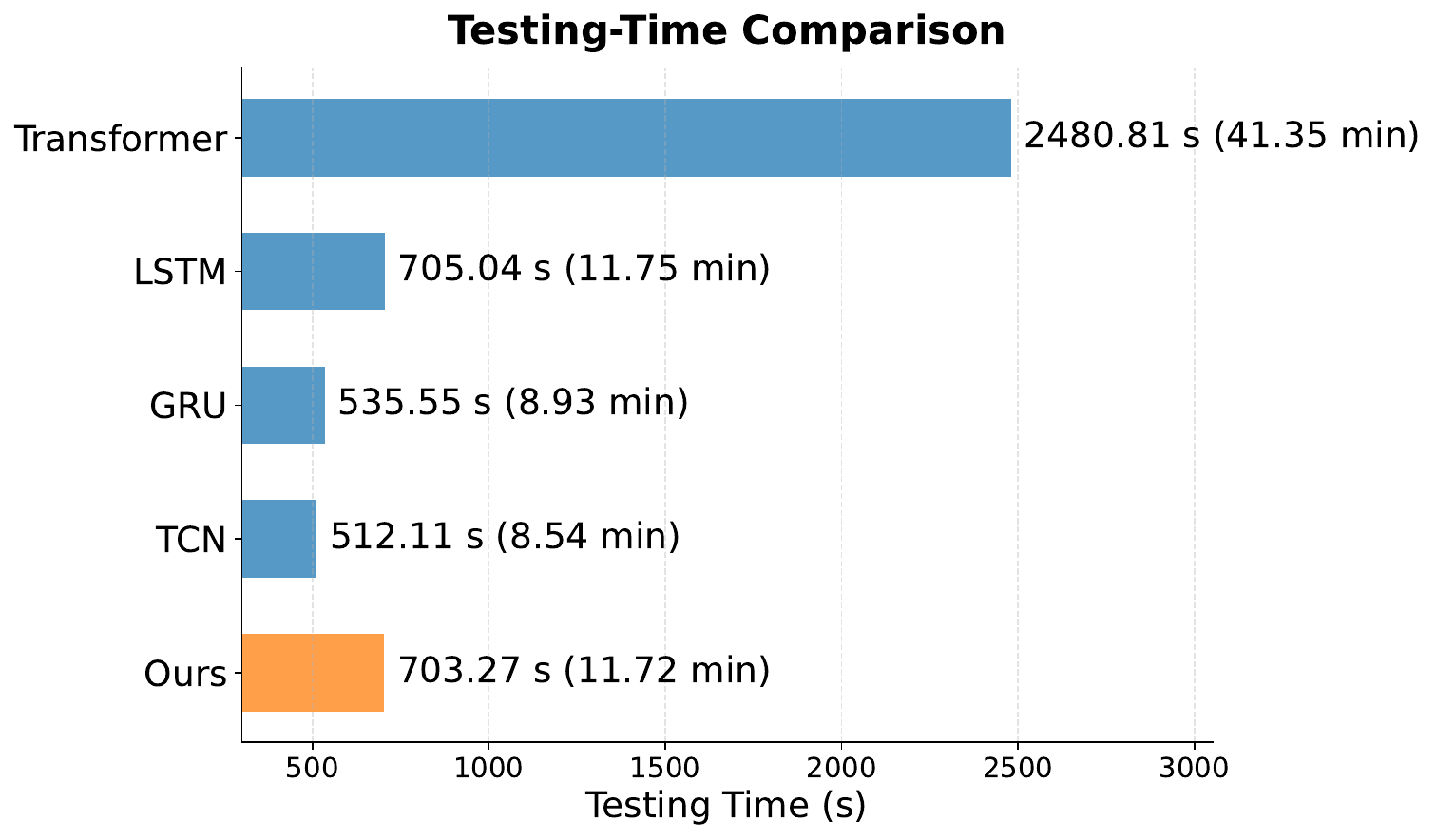}
\caption{Testing time comparison.}
\label{fig:testing_time_comparison}
\end{figure}

Figure~\ref{fig:testing_time_comparison} reports total test-run times for the evaluated configurations. The proposed configuration completes the test in \(703.27~\mathrm{s}\), compared with \(512.11~\mathrm{s}\) for the sensor-only TCN and \(2480.81~\mathrm{s}\) for the Transformer. The configurations perform different preprocessing and estimation steps. These totals describe the measured workstation runs. Embedded viability would require per-sample latency, memory use, and a specified target processor.

\subsection{End-to-End System Results on the Second Dataset}
Table~\ref{tab:final_results_second} reports the original system configurations on LCO\_5 and LCO\_6. The four-input TCN achieved accuracies of \(76.41\%\) and \(73.87\%\), respectively. Its reported F1 scores were \(73.19\%\) and \(70.80\%\). These results quantify performance on the two held-out laboratory cells. Matched-input results are unavailable for this dataset, and the system comparison does not isolate the effect of the classifier.

\begin{table}[H]
\centering
\caption{Original five-stage system results on the held-out LCO cells. The proposed TCN uses sensor inputs and estimated SOC, while the other models use the sensor inputs.}
\label{tab:final_results_second}

\small
\renewcommand{\arraystretch}{1.15}

\resizebox{\textwidth}{!}{%
\begin{tabular}{l c c c c c c c c}
\toprule

\multirow{2}{*}{\textbf{Model}} &
\multicolumn{4}{c}{\textbf{LCO\_5}} &
\multicolumn{4}{c}{\textbf{LCO\_6}} \\

\cmidrule(lr){2-5}
\cmidrule(lr){6-9}

&
\textbf{Acc. (\%)} &
\textbf{Prec. (\%)} &
\textbf{Rec. (\%)} &
\textbf{F1 (\%)} &
\textbf{Acc. (\%)} &
\textbf{Prec. (\%)} &
\textbf{Rec. (\%)} &
\textbf{F1 (\%)} \\

\midrule

Transformer
& 65.5246 \(\pm\) 9.3577 & 68.4855 \(\pm\) 8.3823 & 62.8449 \(\pm\) 7.6372 & 58.3107 \(\pm\) 8.4357 & 65.2170 \(\pm\) 11.0900 & 67.6326 \(\pm\) 11.2158 & 62.3460 \(\pm\) 9.6511 & 58.1104 \(\pm\) 10.0817 \\

LSTM
& 65.9333 \(\pm\) 9.4291 & 65.6130 \(\pm\) 9.7408 & 62.7208 \(\pm\) 9.8691 & 59.5386 \(\pm\) 10.9525 & 66.3952 \(\pm\) 9.0338 & 65.2179 \(\pm\) 11.3722 & 63.1544 \(\pm\) 9.5571 & 59.8775 \(\pm\) 10.7441 \\

GRU
& 66.3954 \(\pm\) 8.8686 & 68.9481 \(\pm\) 8.6038 & 62.9364 \(\pm\) 9.2150 & 58.9520 \(\pm\) 10.2424 & 66.7877 \(\pm\) 8.4203 & 67.4157 \(\pm\) 11.9096 & 63.2802 \(\pm\) 8.8292 & 59.3958 \(\pm\) 9.6772 \\

TCN
& 57.8610 \(\pm\) 8.5378 & 55.4849 \(\pm\) 7.6543 & 55.3820 \(\pm\) 9.1701 & 53.1016 \(\pm\) 10.5778 & 57.9776 \(\pm\) 8.2115 & 55.1710 \(\pm\) 7.8528 & 55.4215 \(\pm\) 8.8417 & 53.4298 \(\pm\) 9.6493 \\

\textbf{Ours}
& \textbf{76.4102 \(\pm\) 8.6691} & \textbf{80.6286 \(\pm\) 6.4054} & \textbf{74.2997 \(\pm\) 7.9076} & \textbf{73.1947 \(\pm\) 9.0215} & \textbf{73.8749 \(\pm\) 6.7874} & \textbf{78.1009 \(\pm\) 7.3395} & \textbf{71.9475 \(\pm\) 5.5979} & \textbf{70.7970 \(\pm\) 6.2824} \\
\bottomrule
\end{tabular}%
}
\end{table}

\section{Discussion}
\label{Discussion}

\paragraph{Relationship to SOC and uncertainty}
Both estimated SOC and the RDS label are expected to decline during complete discharge under these protocols. For an ideal constant-current discharge with \(S_{\mathrm{start}}=1\) and cutoff SOC \(S_{\mathrm{EOD}}=0\), charge balance gives \(S_t=1-(\tau_t-\tau_{\mathrm{start}})/(\tau_{\mathrm{EOD}}-\tau_{\mathrm{start}})=q_t\). The two five-bin labels are then identical. More generally, if SOC declines linearly to a nonzero cutoff SOC, \(q_t=(S_t-S_{\mathrm{EOD}})/(S_{\mathrm{start}}-S_{\mathrm{EOD}})\). Variable current or voltage-limited cutoff can break this relationship. The present experiments do not quantify how often that occurs or whether the additional measured signals materially improve classification beyond estimated SOC alone. A probabilistic interpretation would also require evaluating the calibration of the softmax scores and the stability of the predictions when similar observed histories are followed by different loads.

\paragraph{SOC reference and initialization sensitivity}
A direct assessment would compare \(\widehat S_t\) against the study's SOC label \(S_t^{\rm ref}\) on LCO\_5 and LCO\_6 after documenting the label source and whether it shares current, capacity, or initial-SOC assumptions with the estimator. With SOC expressed as a fraction, define \(e^S_t=100(\widehat S_t-S_t^{\rm ref})\) and report MAE and RMSE in percentage points by cell and temperature. The same complete profiles can be replayed with the nominal initial SOC and deliberately lower initializations (for example, by 10 and 20 percentage points), keeping filter settings and other initial states fixed. A reproducible convergence criterion is the first time the absolute error falls within 2 percentage points and remains there for at least 60 seconds, with trajectories that never meet this criterion reported separately. Because the selected trajectories start fully charged, downward perturbations avoid clipping an attempted positive offset at the upper bound. Comparing the filter with and without neural voltage correction under the same conditions would show how the correction affects voltage fit and SOC recovery. Reference-based errors and convergence were not evaluated in the reported experiments.

\paragraph{Engineering scope}
The available evidence concerns complete single-cell discharges from a fully charged state to a prescribed voltage cutoff. A potential use is progress reporting for a specified battery bench test or for equipment operated under a known duty regime. The stage alone does not specify the minutes remaining, the driving range, or an action threshold. In a vehicle, partial trips, recharging, regenerative braking, changes in demand, and pack-level constraints change the definition and predictability of the endpoint. Field utility would require a defined start and endpoint under such operation, validation on pack or vehicle measurements, and a decision-specific comparison with the existing SOC or remaining-range display. For example, an early-warning application would need to demonstrate useful warning lead time at a measured false-alarm rate. Such evidence is outside the scope of the current laboratory experiments.

\section{Conclusion}
\label{Conclusion}
This study evaluated five-bin normalized remaining discharge time as an ordinal laboratory target. The framework combines a recursive electrothermal SOC estimate with a lightweight temporal classifier whose final-time readout directly accesses three observed samples. The reported five-stage accuracies were \(95.38\%\) and \(81.23\%\) on FTP-75 and PDMHC and \(76.41\%\) and \(73.87\%\) on the two held-out LCO cells. The results establish the observed performance for these complete-discharge protocols. Evidence for added value beyond SOC, robustness to erroneous SOC initialization, and practical value on vehicles requires separate validation.

\clearpage
\FloatBarrier

\section*{Contact Information}
For access to the code and further information about this proposed system, please contact AIWARE Limited Company at: \url{https://aiware.website}

\bibliographystyle{plain}
\bibliography{cas-refs}
\end{document}